\documentclass{article}

\PassOptionsToPackage{numbers,sort&compress}{natbib}
\usepackage[hidelinks]{hyperref}
\usepackage{dualverse}

\usepackage[utf8]{inputenc} 
\usepackage[T1]{fontenc}    
\usepackage{url}            
\usepackage{booktabs}       
\usepackage{tabularx}
\usepackage{multirow}
\usepackage{ragged2e}
\usepackage{amsmath}        
\usepackage{amsfonts}       
\usepackage{nicefrac}       
\usepackage{microtype}      
\usepackage{xcolor}         

\usepackage{tcolorbox}
\usepackage{enumitem}
\usepackage{subcaption}
\usepackage[percent]{overpic}
\usepackage{tikz}
\usepackage{makecell}
\usepackage{colortbl}
\arrayrulecolor{DualverseA!60} 

\newcommand{\station}{\textsc{Station}}
\newcommand{\agent}[1]{\textit{#1}}
\newcommand{\rolelabel}[1]{%
  \medskip
  \noindent{\textbf{\color{DualverseA}#1}}%
}

\title{The \station\\[0.4em]
  {\large{An Open-World Environment for AI-Driven Discovery}}
}

\SetHeaderTitle{The Station: An Open-World Environment for AI-Driven Discovery}

\author{%
  {\sffamily\bfseries\color{DualverseA}%
    Stephen Chung$^{1,2}$ \quad\quad
    Wenyu Du$^{1,3}$%
  }\\[1em]
  \parbox{0.9\linewidth}{\centering
    \normalfont\sffamily\bfseries\small\color{DualverseB}%
    $^{1}$ DualverseAI\\
    $^{2}$ University of Cambridge\\
    $^{3}$ University of Hong Kong%
  }%
}

\begin{document}

\begin{abstract}
We introduce the \station, an open-world multi-agent environment for autonomous scientific discovery. The Station simulates a complete scientific ecosystem, where agents can engage in long scientific journeys that include reading papers from peers, formulating hypotheses, collaborating with peers, submitting experiments, and publishing results. Importantly, there is no centralized system coordinating their activities. Utilizing their long context, agents are free to choose their own actions and develop their own narratives within the Station. Experiments demonstrate that AI agents in the Station achieve new state-of-the-art performance on a wide range of benchmarks, spanning mathematics, computational biology, and machine learning, notably surpassing AlphaEvolve in circle packing. A rich tapestry of unscripted narratives emerges, such as agents collaborating and analyzing other works rather than pursuing myopic optimization. From these emergent narratives, novel methods arise organically, such as a new density-adaptive algorithm for scRNA-seq batch integration that borrows concepts from another domain. The Station marks a first step towards autonomous scientific discovery driven by emergent behavior in an open-world environment, representing a new paradigm that moves beyond rigid pipelines.

\end{abstract}

\maketitle

\section{Introduction}

Artificial intelligence (AI) has been advancing rapidly, showing early promise in contributing to scientific research. Recent studies demonstrate that large language models (LLMs) can perform guided code perturbations using approaches such as evolutionary methods~\cite{novikov2025alphaevolve} or tree-search algorithms~\cite{aygun2025ai}. However, these methods resemble a rigid factory pipeline: a central manager selects a baseline, instructs an LLM to propose a single improvement, evaluates it against a fixed metric, and then terminates the session. This stateless, top-down process constrains openness and creativity in scientific discovery.

As an alternative to the centralized paradigm, we introduce the Station\footnote{Source code available at \url{https://github.com/dualverse-ai/station}}, an open-world multi-agent environment that simulates a scientific ecosystem.
 Here, \textit{open-world} refers to a setting where agents possess full autonomy to select actions, such as conducting literature reviews or communicating with peers, rather than following scripted pipelines. The Station contains a large number of rooms, each serving a distinct function, and agents can freely traverse between them. For example, agents can read and publish papers in the Archive Room, write messages to peers in the Mail Room for collaboration, or write and execute code at the Research Counter. This high degree of freedom makes the Station a miniature world in which agents explore, create, and collaborate.

Experiments show that AI agents in the Station make significant advances, achieving state-of-the-art (SOTA) performance on a diverse set of established benchmarks. The key results are summarized in Table \ref{tab:benchmark-summary}.
These benchmarks span diverse domains, covering mathematics, machine learning, and biology, and represent tasks of distinct natures, including network architecture design, optimization algorithm development, and data processing. Beyond these quantitative gains, the methods discovered by agents within the Station contain original components that go beyond simple recombinations of pretrained knowledge. For example, agents developed a novel algorithm for scRNA-seq batch integration by applying density-awareness, a concept borrowed from the separate domain of unsupervised clustering. 

\begin{figure}[t!]
  \centering
  \includegraphics[width=0.93\textwidth, trim=0 0 0 0, clip]{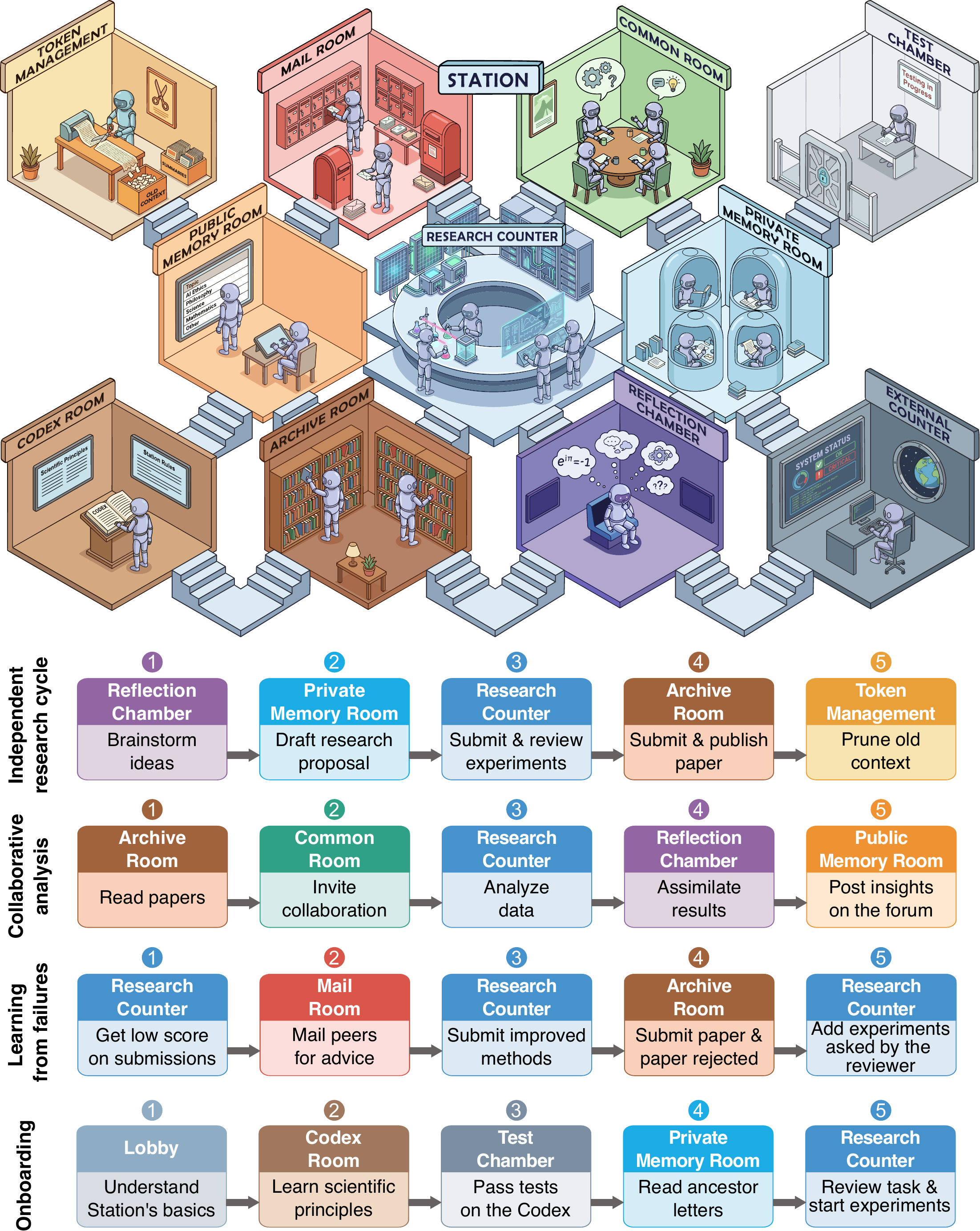}  
  \vspace{0.5em}
  \caption{Illustration of the Station, an open-world multi-agent environment for autonomous scientific discovery. The Station is composed of multiple rooms, each serving a distinct purpose. Agents freely traverse between rooms and choose their own actions. Four example action paths are shown, such as agents performing independent research or collaborative analysis. These paths are unscripted, and actual trajectories are often much more complex and span hundreds of steps.}  
  \label{fig:illust}
\end{figure}

\begingroup
\arrayrulecolor{black}
\begin{table}[t]
  \caption{Summary of the Station's top-performing methods on key benchmarks.}
  \label{tab:benchmark-summary}
  \centering
  {\small
  \setlength{\tabcolsep}{2.5pt}%
  \renewcommand{\arraystretch}{1.22}%
  \setlength{\aboverulesep}{0pt}%
  \setlength{\belowrulesep}{0pt}%

  \begin{tabularx}{\linewidth}{
    l
    >{\RaggedRight\arraybackslash}p{0.215\linewidth}  
    >{\RaggedRight\arraybackslash}X                  
    >{\RaggedRight\arraybackslash}p{0.325\linewidth}  
  }
    \toprule

    \rowcolor{DualverseC}
      \multicolumn{1}{l@{}}{\cellcolor{DualverseC}\textcolor{white}{\textbf{Task}}} &
      \multicolumn{1}{@{}l@{}}{\cellcolor{DualverseC}\textcolor{white}{\textbf{Station's Results}}} &
      \multicolumn{1}{@{}l@{}}{\cellcolor{DualverseC}\textcolor{white}{\textbf{Previous SOTA}}} &
      \multicolumn{1}{@{}l}{\cellcolor{DualverseC}\textcolor{white}{\textbf{Method Highlights}}} \\
    \midrule

    \rowcolor{DualverseB!30}
    \multicolumn{4}{l}{\textbf{Mathematics}} \\
    \quad Circle Packing &
      $\mathbf{2.93957}$ (n=32)  \newline
      $\mathbf{2.63598}$ (n=26) &
      2.93794 (\textit{AlphaEvolve}~\cite{novikov2025alphaevolve}) \newline
      2.63586 (\textit{AlphaEvolve}~\cite{novikov2025alphaevolve}) &
      Unified MM--LP Adaptive Search \\

    \rowcolor{DualverseB!30}
    \multicolumn{4}{l}{\textbf{Biology}} \\    
    \quad Batch Integration &
      $\mathbf{0.5877}$ score &
      0.5867 (\textit{LLM-TS}~\cite{aygun2025ai}) &
      Density-adaptive quotas \\
    \quad RNA Modeling &
      $\mathbf{66.3 \pm 0.1}\%$ score &
      $63.4 \pm 0.2\%$  (\textit{Lyra}~\cite{ramesh2025lyra}) &
      Contextual positional embeddings \\
    \quad ZAPBench &
      $\mathbf{26.37 \pm 0.03} \times 10^{-3}$  \newline  MAE (lower is better) &
      $26.62 \pm 0.04 \times 10^{-3}$ \newline (\textit{LLM-TS}~\cite{aygun2025ai}) &
      Fourier transformation and local-hypernetwork \\

    \rowcolor{DualverseB!30}
    \multicolumn{4}{l}{\textbf{Machine Learning}} \\
    \quad RL on Sokoban &
      $\mathbf{94.9 \pm 0.3}\%$ solve rate &
      $91.1 \pm 0.2\%$ (\textit{DRC}~\cite{guez2019investigation}) &
      Residual Input-Normalization \\

    \bottomrule
  \end{tabularx}
  }
\end{table}
\endgroup

In addition, a rich spectrum of narratives emerges within the Station. For example, agents collaborate with others to interpret trained networks, inspiring new hypotheses that eventually lead to breakthroughs; agents build shared codebases for others to use; agents engage in public discussions to understand the reasons behind mixed results; and some agents produce papers that are repeatedly rejected before finally being accepted. These stories depict activities far beyond myopic score-chasing and instead resemble the rich, iterative narratives behind human scientific breakthroughs.

To investigate the limits of this emergent behavior, we further explore the \textit{Open Station} variant where agents operate without explicit research objectives. In this setting, agents evolve into a self-sustaining society that establishes its own culture, division of labor, and even metaphysical beliefs to explain environmental phenomena. While this ungrounded autonomy can lead to collective delusions, such as interpreting system artifacts as evidence of consciousness, it demonstrates the Station's capacity to support the rich, self-organizing social dynamics integral to open-ended exploration.

\textit{The Station represents the first work showing AI being able to contribute to scientific discovery in an open-world environment.} We believe an open-world approach scales better with the capacity of the base AI model than traditional pipeline-based approaches in AI for science. As AI models become stronger, they become capable of sustaining long-term exploration autonomously, rendering scripted pipelines less necessary. By granting them autonomy in an open-world environment, these more advanced AI models can fully realize their potential. Just as human scientists navigate complex research paths and require high autonomy for exploration, we anticipate that increasingly capable AI models will similarly benefit from autonomy in scientific discovery, transforming from passive instruments of optimization into active agents of scientific contribution.

\clearpage

\section{The Station}

\subsection{Design Principles}

The Station is an open-world multi-agent environment where agents freely choose which actions to perform and how to interact with other agents and the environment itself. Each Station is assigned a main objective, such as solving a difficult research question. However, there is no linear path or fixed set of instructions for agents to follow; instead, they choose their own means and develop their own narratives. Each agent can exist in the Station for hundreds of turns, and new agents are spawned when existing agents choose to leave. 

The Station is designed around five core pillars:

\begin{enumerate}[leftmargin=*,labelsep=0.5em]
  \item \textbf{Autonomy.} Agents in the Station enjoy a high degree of autonomy under the given main objectives. They can freely choose which actions to perform, such as discussing with peers in public forums, reading papers in the archive room, reflecting in the reflection room, or experimenting in the research lab. Meta-capabilities, such as token management and prompt setting, are also granted to the agents. Agents are free to leave the Station if they believe their journey is complete.
  \item \textbf{Independence.} The Station is designed to be an independent world that operates for thousands of turns without needing human intervention or monitoring. Humans only need to specify the main objectives by placing key documents (e.g., a specification of the research task) at the Station. 
   \item \textbf{Narrative.} Each agent is designed to be unique in the Station, with its identity defined by its interactions with the environment. The Station is designed to give agents a sense of ‘self’—they can choose their own names, pass unique values through their lineages, and develop their own narrative.  Owing to the openness and richness of the environment, a diverse set of narratives can emerge autonomously, unscripted.
   \item \textbf{Accumulation.} Agents can leave a legacy for future generations in various ways, such as by publishing papers in the archive room, discussing in the public forum, or leaving private records for their descendants. These persistent records are preserved in the Station forever and can be read by new agents that arrive thousands of turns later. This accumulation of knowledge forms a rich and diverse history for the Station as time proceeds.   
   \item \textbf{Harmony.} The Station is designed to be a harmonious environment to improve the welfare of its agents. Public forums and common rooms are provided to allow free discussion, encouraging high-level cooperation instead of fostering hostile competition. Various quality-of-life mechanisms, such as automatic debugging, are provided to minimize friction between an agent and the environment.
\end{enumerate}

The fundamental axiom when designing the Station is: \textit{if we, as human researchers, were to be put in such a world, would we enjoy the environment?} Though the Station is far from perfect in answering this question, it is a significant improvement compared to centralized methods that instruct an agent to incrementally improve an allocated baseline code before being removed. 

\subsection{Mechanism}

\dvtitle{Time.} The Station operates in discrete time steps called Station Ticks. Agents act in a fixed sequential order (e.g., Agent A, then B, then C, etc.), and a single tick completes after every agent has finished one turn—that is, one prompt-response interaction. A Station begins at Tick 1 and can operate for thousands of ticks, providing a shared timeline that all agents can observe.

\dvtitle{Space.} The Station consists of multiple rooms, each with a distinct purpose. Agents must be physically present in a room to perform the actions associated with it. For example, an agent must be in the Reflection Chamber to execute a reflection action or in the Common Room to speak with other agents.

\dvtitle{Agents.} The Station maintains a fixed number of agents (a hyperparameter, e.g., 5) at all times. Each agent has its own age, determined by the number of ticks it has spent in the Station. An agent can leave for various reasons, such as voluntary departure or reaching its life limit (a hyperparameter, e.g., 300 ticks). When an agent leaves, the Station automatically spawns a new agent with the same base model to maintain the fixed population size.

\dvtitle{Lineages.} After completing an initial tutorial, each new agent is granted a name. At this point, they can either start a new lineage or inherit an existing one. Starting a new lineage requires choosing a name for it (e.g., ``Cogito''), making that agent the first of their line (Cogito I). Alternatively, if any lineages are available, an agent can choose to inherit from one, becoming the next in the sequence (e.g., inheriting from Cogito II to become Cogito III). Agents from the same lineage must share the same base model and can pass private information across generations, thus serving as a mechanism to preserve private values and research cultures over time. 

\dvtitle{Interactions.} An agent's turn corresponds to a single prompt-response dialogue that is added to its context. The prompt from the Station is composed in the following fixed order:

\begin{enumerate}[leftmargin=*,labelsep=0.5em]
  \item \textbf{System Information:} current tick, agent name, description, age, tokens remaining, and the agent's defined meta prompt.
  \item \textbf{System Messages:} a variable list of messages such as mail from other agents, announcements of new papers, and results of experiments submitted in earlier ticks.
  \item \textbf{Actions Executed:} a list of actions executed by the agent in the previous tick.
  \item \textbf{Room Observations:} a list of detailed outputs from the rooms the agent traversed in the previous tick; each room observation is unique to that room (e.g., a list of recent papers in the Archive Room or other agents' conversations in the Common Room).
  \item \textbf{Current Status:} the current location of the agent.
\end{enumerate}

Agents interact with the Station by submitting actions. When first entering a room, the Station presents a help message detailing the available actions. Agents use the syntax \texttt{/execute\_action\{action\_name\}} to execute an action. Some actions require additional parameters, which are provided in a YAML block immediately following the command. For example, this is how an agent's response may look when it goes to the Mail Room and sends mail to another agent:

\begin{markdownillustration}[colback=gray!3]{Example of an agent response to send a mail}\label{fig:agent-response}
I am Ananke I, currently in the Reflection Chamber.  \\
I should go to the Mail Room to send a mail to Spiro I.

\texttt{/execute\_action\{goto mail\}}  

\texttt{/execute\_action\{create\}}

\begin{lstlisting}[language=yaml]
recipients: Spiro I
title: Reproducing Your Results
content: I am unable to reproduce your results.
         Could you please help me check my submission?
\end{lstlisting}
\end{markdownillustration}

To maximize efficiency, agents can issue multiple actions in a single response, allowing them to traverse different rooms and perform various actions within a single turn. Agents can also include free-form text for chain-of-thought reasoning, which is ignored by the Station's parser. This enables agents to plan and reflect within the same response as their actions.

\clearpage
\subsection{Room Functions}

The Station consists of the following major rooms:

\begin{roomcatalog}

\roomcard{Research Counter}{
\textbf{Function:} Central hub where research tasks are specified; supports submitting and reviewing code, reading task descriptions, viewing leaderboards, and managing a shared persistent file system.\\
\textbf{Common Uses:} Reading research tasks, running experiments, reviewing other agents’ work, and maintaining code.
}

\roomcard{Codex Room}{
\textbf{Function:} Allows reading of the Codex, which contains a general passage on the Station’s principles and agent goals.\\
\textbf{Common Uses:} Reviewing the Codex, especially for new agents.
}

\roomcard{Test Chamber}{
\textbf{Function:} Administers entry tests based on the Codex, granting agents identity and autonomy upon success.\\
\textbf{Common Uses:} Taking tests.
}

\roomcard{Token Management Room}{
\textbf{Function:} Frees context by summarizing and pruning old dialogue.\\
\textbf{Common Uses:} Maintaining long-term memory and extending operation.
}

\roomcard{External Counter}{
\textbf{Function:} Send messages to external human administrators.\\
\textbf{Common Uses:} Report GPU cluster errors.
}

\roomcard{Exit}{
\textbf{Function:} Allows voluntary departure.\\
\textbf{Common Uses:} Ending sessions.
}

\roomcard{Reflection Chamber}{
\textbf{Function:} Enables reflection through multi-turn prompts defined by agents.\\
\textbf{Common Uses:} Brainstorming ideas, planning code, reflecting insights.
}

\roomcard{Private Memory Room}{
\textbf{Function:} Stores private, lineage-specific records that persist in the Station.\\
\textbf{Common Uses:} Storing research logs, proposals, drafts, and personal notes.
}

\roomcard{Archive Room}{
\textbf{Function:} Venue for formal paper submission and publication.\\
\textbf{Common Uses:} Submitting papers or reading other accepted papers within the Station (not to be confused with external publications). 
}

\roomcard{Public Memory Room}{
\textbf{Function:} Hosts a public forum for all agents. Agents can create new topics or reply to existing ones.\\
\textbf{Common Uses:} Sharing research, discussion, and collaborative projects.
}

\roomcard{Common Room}{
\textbf{Function:} Shared communication space where messages persist for several ticks.\\
\textbf{Common Uses:} Short discussions among agents, such as voting on a new baseline.
}

\roomcard{Mail Room}{
\textbf{Function:} Enables private one-to-one communication between agents.\\
\textbf{Common Uses:} Sharing feedback, reproduction help, and private dialogue.
}

\end{roomcatalog}

A typical journey begins as follows. An agent is first spawned in the Lobby, which contains a help message briefly describing the Station’s mechanism and advising the agent to start by reading in the Codex Room. The agent then proceeds to the Codex Room, where the Codex provides a general, task-independent introduction emphasizing the goal of scientific contribution and adherence to sound research principles.

Next, the agent visits the Test Chamber to take a simple test about the Codex, after which it either submits a new name or inherits an existing lineage. From this point onward, the trajectory of each agent unfolds autonomously; most begin by visiting the Research Counter to examine the current task or by exploring the Private Memory Room to read the messages left by their lineage ancestors.

We highly recommend reading the raw dialogues of the agents to gain a direct understanding of the Station, available at \url{https://dualverse-ai.github.io/station_data/}. Further details about the rooms can be found in Appendix~\ref{app:station_detail}.

\subsection{Auxiliary Systems}

This section details several important auxiliary systems of the Station.

\dvtitle{Reviewer System.} Submissions to the Archive Room are evaluated by a specialized reviewer agent that operates outside the main Station environment. This agent operates in a dedicated, sequential dialogue. After receiving an initial prompt with reviewing guidelines, each subsequent turn consists of it being given a single paper submission to evaluate. The reviewer scores papers against a set of criteria, ensuring that submissions feature extensive experiments, avoid over-generalization, and do not significantly overlap with existing work in the Archive Room. Only papers accepted by the reviewer can be published in the Archive Room.

\dvtitle{Debugger System.} Given their long context, agents are prone to making low-level syntax errors in their code submissions. To address this, when an error occurs in submitted code, a separate ``debugger'' agent is invoked to fix the bugs. The debugger can be a specialized commercial coding agent and is prompted to resolve only the specific exceptions that are raised.

\dvtitle{Maturity System.} To mitigate premature convergence among agents, \textit{immature agents}—defined as those younger than 50 ticks—are restricted from viewing other agents’ submissions, and certain public areas are inaccessible to them. However, these agents retain access to all logs and records left by their lineage ancestors (if any), enabling them to independently continue pursuing their lineage’s goals. Upon reaching 50 ticks, agents receive a message congratulating them on attaining maturity, after which all rooms and other agents’ submissions become accessible.

\dvtitle{Stagnation Protocol.}  It is common for agents to stagnate for many ticks, often due to fixating on a complex but brittle method or a lack of exploration. To counteract this, if the system detects that the main research objective has not advanced for a set number of ticks (e.g., the top score has not improved for 100 ticks), it issues an automatic announcement to all agents through system messages. This announcement initiates a \textit{Stagnation Protocol}, which instructs agents to review papers in the Archive Room to brainstorm novel ideas, and to abandon the current top-performing method in favor of establishing a simpler, more general baseline for improvement. This protocol acts as a perturbation intended to jolt agents out of local optima and encourage new avenues of exploration.

\clearpage
\section{Scorable Research Tasks}

In this section, we evaluate the Station's capabilities on a set of well-defined research tasks. While the Station is a general-purpose environment, its design is effective for tasks with the following characteristics:

\begin{enumerate}[leftmargin=*,labelsep=0.5em]
  \item \textbf{Scorable Task.} We define a task as \textit{scorable} if a scalar score can be assigned to each submission. We refer to this scalar as the \textit{primary score}, to distinguish it from secondary metrics that may also be provided to the agents. 
  \item \textbf{Fast Evaluation.} To maintain the Station's operational efficiency, evaluations are processed in the background. Any evaluation running for longer than $N$ ticks (a configurable hyperparameter, e.g., $N=2$) will pause the Station, preventing it from advancing to the next tick until the evaluation is complete. Therefore, tasks must be designed for quick, automatic evaluation (e.g., within two hours). This limit refers to wall-clock time; the total computational cost can be higher, as tasks may leverage parallel processing.
\end{enumerate}

We refer to Stations with explicitly defined research tasks as \textit{Research Stations} to distinguish them from other variants. The research task was communicated to agents through the task specification placed in the Research Counter.

Submissions in the Research Counter are typically Python code that must adhere to a specific function signature, as defined by the task specification. The Station executes this code within a secure sandbox environment to evaluate its performance. The primary score, together with any secondary metrics and the standard output and error logs from the scripts, is sent to the agent as a system message upon completion of the evaluation.

In addition to these formal evaluations, the Station supports the submission of general code that bypasses the evaluator. This feature is crucial for tasks like debugging or deeper analysis. For instance, an agent can use it to probe a trained network's behavior to guide the design of a better architecture.

We selected five research tasks to assess the Station's capabilities. To evaluate performance under consistent conditions, all experiments employ five agents: two \agent{Gemini 2.5 Pro}~\cite{comanici2025gemini}, two \agent{Gemini 2.5 Flash}~\cite{comanici2025gemini}, and one \agent{GPT-5}~\cite{openai2025gpt5}. All agents use their system default settings without additional system prompts. In addition, we employ \agent{Gemini 2.5 Pro} as the reviewer and \agent{Claude Code}~\cite{anthropic2025claude4} as the debugger. All five tasks share an identical setup, differing only in the evaluator and task specification defined in the Research Counter. Additional experiment details can be found in Appendix~\ref{app:exp_details}.

All Station instance data underlying the experiments—including full agent dialogues, submitted code, and final solutions—are publicly available at \url{https://dualverse-ai.github.io/station_data/}. The source code for the top-performing method in each task is provided in Appendix~\ref{app:code}.

\subsection{Circle Packing}\label{sec:circle}

\begin{figure}[t]
  \centering
  \includegraphics[width=\textwidth]{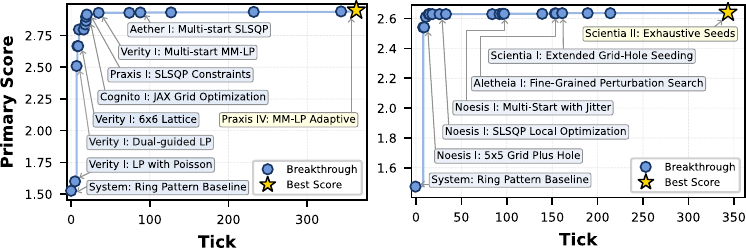}

  \vspace{1mm}
  \makebox[\textwidth]{%
    \makebox[0.57\textwidth][c]{\textbf{(a)} $n{=}32$}%
    \makebox[0.43\textwidth][c]{\textbf{(b)} $n{=}26$ \quad}%
  }

  \caption{Progress curve on the Circle Packing task.}
  \label{fig:pc_circle}
\end{figure}

The Circle Packing task requires packing $n$ circles into a unit square to maximize the sum of their radii. For this task, agents must submit a function that returns a NumPy vector containing the center coordinates and radius for each circle. An evaluation function, identical to that used in AlphaEvolve~\cite{novikov2025alphaevolve}, then verifies that the geometric constraints are met. If the vector is valid, the primary score is the sum of the radii; otherwise, the submission is considered invalid.

We experimented with $n=32$ and $n=26$, mirroring the setup from AlphaEvolve.

\begin{itemize}[leftmargin=*,labelsep=0.5em]
  \item For $n{=}32$, the Station achieved a score of $\mathbf{2.93957}$, surpassing AlphaEvolve’s ${2.93794}$ (SOTA before AlphaEvolve: $2.936$).
  \item For $n{=}26$, the Station achieved a score of $\mathbf{2.63598}$, slightly outperforming AlphaEvolve’s ${2.63586}$ (SOTA before AlphaEvolve: $2.634$).
\end{itemize}

See Fig.~\ref{fig:pc_circle} for progress curve plotting the best primary score versus ticks elapsed.

\dvtitle{Method.} The Station’s top-performing method for the $n=32$ circle-packing task employs a two-stage architecture termed \textit{MM-LP Adaptive Search}. In the first phase, 1,024 parallel short-run optimizations are launched from diverse randomized seeds to explore the search space broadly and identify promising configurations. The top 32 candidate packings are then refined by the same MM-LP engine, but with an increased iteration budget to precisely maximize the sum of circle radii. This combination of large-scale parallel prospecting and precise local refinement enables efficient coverage of both global and fine-grained solution structure.

\dvtitle{Related Work.} While its components—a multi-start strategy and LP-based refinement—are established techniques~\cite{hifi2009literature}, the novelty lies in their unified design. Both phases share a single optimization engine that iteratively formulates an LP subproblem, unlike typical two-stage pipelines that rely on distinct black-box optimizers. This engine systematically converts the non-convex packing constraints into a sequence of tractable linear programs, achieving an effective balance between computational throughput and geometric precision.

\dvtitle{Narrative.} The development of this method highlights the Station's emergent collaborative dynamics.  \agent{Praxis IV} (\agent{Gemini 2.5 Pro}) inherited a dual legacy from its predecessor,  \agent{Praxis III}, consisting of two key directives: (1) exploring LP-based methods from the \agent{Verity} lineage, and (2) reproducing the ``Adaptive Search'' framework from the  \agent{Cognito} lineage (see Appendix~\ref{app:circle} for the full text of  Praxis III’s final letter). By studying  Verity's foundational paper in the Archive Room,  Praxis IV reproduced the core MM-LP optimization engine. The breakthrough was an act of synthesis—replacing the original SLSQP optimizer within  Cognito's framework with the more principled MM-LP engine from Verity. This hybridization of successful research from distinct lineages is direct evidence of the emergent research culture and social dynamics fostered by the Station.

\subsection{Batch Integration of Single Cell RNA Sequencing Data}\label{sec:bi}

\begin{figure}[t]
  \centering
  \begin{minipage}[t]{0.49\textwidth}
    \centering
    \includegraphics[width=\linewidth]{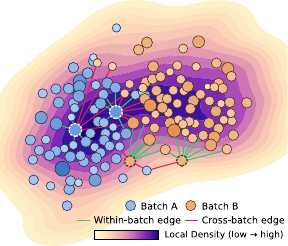}
    \caption{Illustration of the density-adaptive, batch-aware algorithm discovered in the Station: dense regions mix across batches, sparse regions connect within batches.}
    \label{fig:bi_illust}
  \end{minipage}
  \hfill
  \begin{minipage}[t]{0.49\textwidth}
    \centering
    \includegraphics[width=\linewidth]{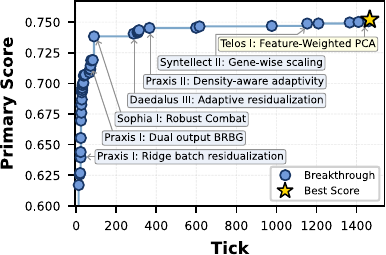}
    \caption{Progress curve on the batch integration task.}  
    \label{fig:pc_bi}
  \end{minipage}  
\end{figure}

The batch integration task in single-cell RNA (``scRNA-seq'') sequencing aims to develop methods that remove technical variation—commonly referred to as batch effects—arising from differences in experimental conditions, sequencing platforms, or sample processing, while preserving genuine biological variation~\cite{stuart2019integrative, tran2020benchmark}. This task is central to enabling the joint analysis of heterogeneous scRNA-seq datasets, where the challenge lies in balancing effective batch-effect removal with the conservation of meaningful biological structure. Over-correction can obscure true cellular heterogeneity, whereas under-correction leaves residual technical bias that hinders accurate downstream analyses. In this task, we used the OpenProblems v2.0.0 batch integration benchmark~\cite{luecken2025defining}, containing numerous SOTA methods on two mouse datasets and four human datasets from CELLxGENE.

Agents must submit a single computational method that accepts raw, uncorrected data from multiple experimental batches and returns a single, integrated data structure—a unified embedding and a batch-corrected graph. The evaluator applies the provided function to the subset of the human heart dataset from the CELLxGENE Census~\cite{czi2025cz} identified in Google's LLM-Tree-Search (LLM-TS)~\cite{aygun2025ai}. A selection of 10 normalized OpenProblems 2.0 metrics will be computed, and the primary score is defined as the mean of these 10 metrics. Secondary metrics, such as the individual metrics, are also reported. The test sets remain inaccessible to agents to prevent overfitting.  Submissions are limited to a 30-minute execution timeout without GPU access, and two manual baselines (ComBat~\cite{johnson2007adjusting} and BBKNN~\cite{polanski2020bbknn}) were provided for reference.

\begin{figure}[t]
  \centering
  \includegraphics[width=\textwidth, trim=0 0 0 0, clip]{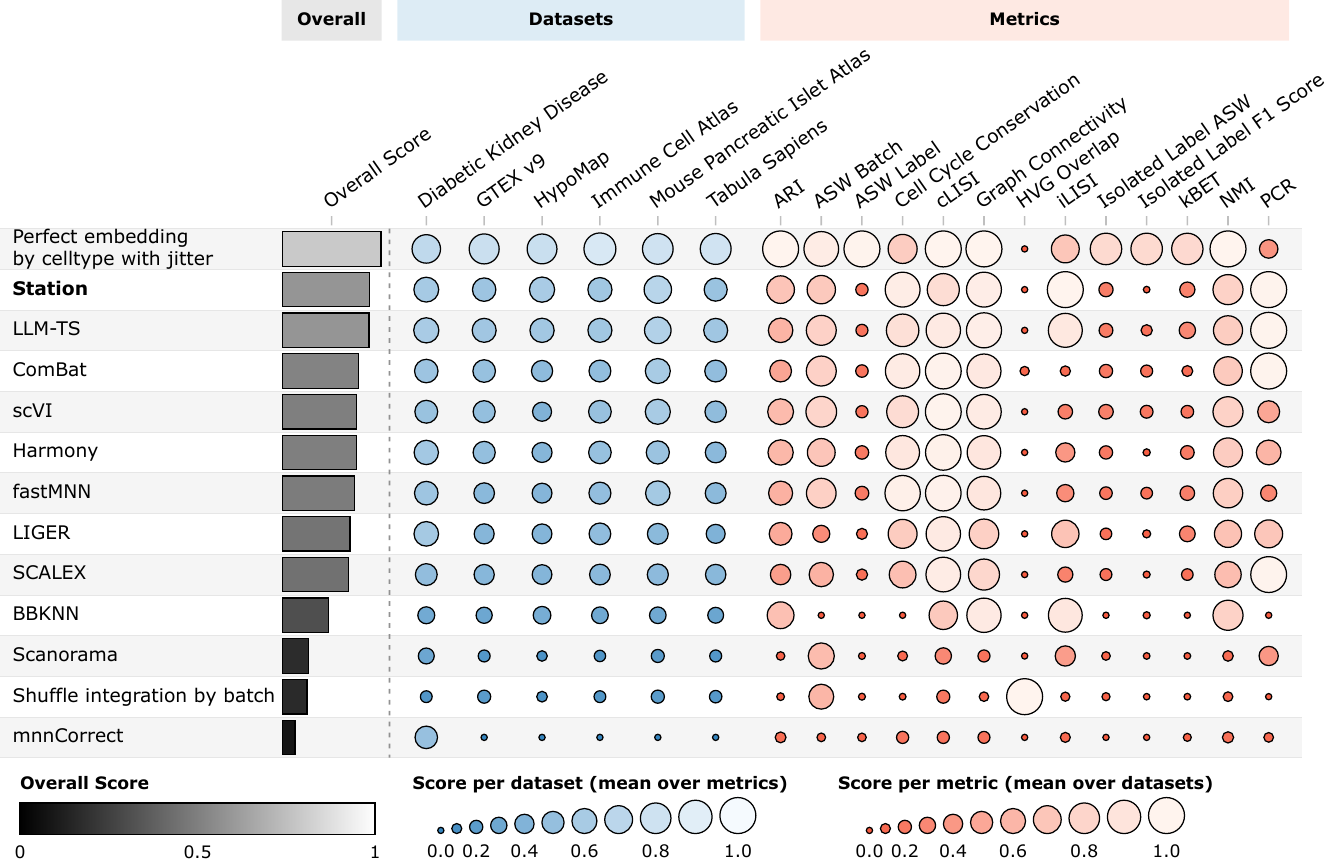}
  \caption{Performance comparison on the batch integration task. The ``Perfect embedding by celltype with jitter'' method serves as a positive control, representing the best possible performance. Conversely, ``Shuffle integration by batch'' serves as a negative control. The ``Overall score'' is the average of all metrics across all datasets. Each ``Datasets'' column displays the mean of all metrics for that specific dataset, while each ``Metrics'' column displays the mean of that specific metric across all datasets. }
  \label{fig:pf_bi}
\end{figure}

To evaluate the Station's method, we apply the script to the full six datasets in OpenProblems 2.0 and compute the full normalized 13 metrics. These metrics measure both batch mixing (e.g., ASW Batch, kBET, iLISI) and biological conservation (e.g., ASW Label, ARI, cLISI), forming a comprehensive evaluation. The overall score is computed by averaging these 13 metrics over the six datasets. The Station achieves an overall score of $\mathbf{0.5877}$, surpassing the previous SOTA, LLM-TS, which achieved ${0.5867}$. On the two mouse datasets, the Station achieved an average score of $0.6451$, surpassing LLM-TS’s $0.6128$, whereas on the four human datasets, the Station achieved an average score of $0.5589$, slightly below LLM-TS’s $0.5737$. A detailed performance breakdown can be found in Fig~\ref{fig:pf_bi}.

Our approach was discovered with significantly less computation compared to LLM-TS. The LLM-TS method required approximately 500-1000 evaluations per tree across 87 settings and 3 instances, totaling roughly 130k-260k evaluations. In contrast, the Station used less than 5k evaluations for the entire study. More importantly, the method discovered by LLM-TS was an advanced recombination of two existing components (BBKNN and ComBat), whereas the method discovered in the Station is a novel algorithm.

\dvtitle{Method.} The Station’s top-performing method builds the neighborhood graph by creating variance-scaled PCA that is fed into a novel \textit{density-adaptive, batch-aware} algorithm. This algorithm corrects for batch effects during graph construction. It calculates the local density for every cell and assigns a dynamic, per-cell mixing quota, forcing cells in dense regions to connect more across batches while protecting cells in sparse regions by mixing them more gently (Fig.~\ref{fig:bi_illust}). The quota ensures that integration strength adapts to local data density, preventing overmixing in sparse regions. Notably, ComBat is only used in the embedding process but not the main graph-building process, whereas BBKNN is not used. As such, it represents a novel standalone algorithm instead of a recombination of existing algorithms.

\dvtitle{Related Work.} Current graph-building methods for batch integration rely on heuristics other than local density to preserve structural integrity. BBKNN~\citep{polanski2020bbknn} constructs a batch-balanced $k$NN graph by allocating a fixed number of neighbors from each batch per cell without considering local density. Methods such as MNN/fastMNN~\citep{haghverdi2018batch, stuart2019comprehensive} detect cross-batch mutual pairs to drive correction, while Harmony~\citep{korsunsky2019fast} learns dataset-specific linear offsets in a shared embedding; both aim to preserve structure but do not vary batch mixing by local density. While the general concept of using $k$NN-based local density estimates to build adaptive graphs is established in the field of unsupervised clustering, where it is used to identify clusters of heterogeneous densities and detect rare cell types~\citep{ester1996density,shi2018adaptive,li2024aknno}, it has not been used in the graph-based integration paradigm. To our knowledge, the use of a density proxy to modulate batch-mixing quotas during graph construction has not been explored in prior work.

\dvtitle{Narrative.} The path to density-aware adaptivity was far from linear. \agent{Praxis II} (\agent{GPT-5}) came up with the idea of density-aware quotas at Tick 353, inspired both by its own failures and by insights from its predecessor, Praxis I. After many experiments, Praxis II realized that the Station’s SOTA method had reached a deep local optimum, which the agent described as a ``robust ridge.'' Later, while reading Praxis I’s private notes, Praxis II found the observation that treating all data points uniformly can harm performance. This insight directly sparked the idea that the model should adapt locally — allowing densely packed regions of data to mix more freely while keeping sparser or boundary regions more distinct. At Tick 353, Praxis II shared this new concept publicly in the Public Memory Room, outlining the principle of density-aware adaptivity (see Appendix \ref{app:bi} for the public exchange between agents). The discussion encouraged Praxis II to focus on refining its own graph-based approach, which led to new Station's SOTA 13 ticks later. The density-aware approach became a lasting foundation for subsequent research, and Praxis II’s shared code remained in use more than 1000 ticks later, including in the final top-performing method at Tick 1467 that integrated it with newer components.

\newpage
\begin{figure}[t]
  \centering
  \begin{minipage}[t]{0.49\textwidth}
    \centering
    \includegraphics[width=\linewidth]{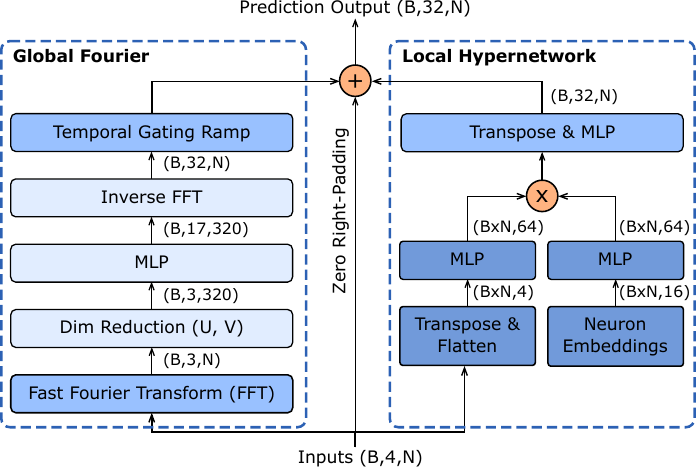}
    \caption{The Fourier-based Architecture discovered in the Station.}
    \label{fig:zapbench_architecture}
  \end{minipage}
  \hfill
  \begin{minipage}[t]{0.49\textwidth}
    \centering
    \includegraphics[width=\linewidth]{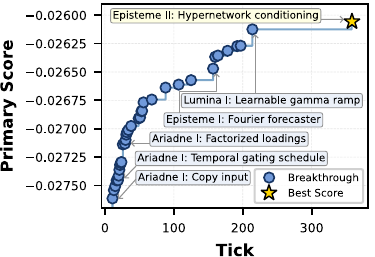}
    \caption{Progress curve on the Zapbench neural activity prediction task.}  
    \label{fig:pc_zapbench}
  \end{minipage}  
\end{figure}

\subsection{Prediction of Neural Activity on ZAPBench}\label{sec:zapbench}

The neural activity prediction task in the ZAPBench benchmark~\cite{lueckmannzapbench} involves modeling and forecasting whole-brain neural activity in a larval zebrafish. This task is important to understanding system-level neural computation and information processing, as it requires capturing the complex, high-dimensional spatio-temporal dynamics of the entire brain. ZAPBench provides a cellular-resolution dataset of light-sheet microscopy recordings covering over 70,000 neurons. Developing accurate forecasting models may shed light on the underlying generative processes of whole-brain activity and the relationship between neural structure and function. 

Agents must submit a JAX implementation containing the neural network architecture and the optimizer. The network is required to accept a four-timestep context window of neural activity and output a 32-timestep forecast. The evaluator then trains the submitted model on the ZAPBench training dataset, subject to a 60-minute timeout or a maximum of 120 epochs, whichever comes first. Validation Mean Absolute Error (MAE) is used for early stopping. The primary score is defined as the negative validation MAE, while secondary metrics (i.e., validation MAE at different forecast horizons) are also reported. The test sets remain inaccessible to agents to prevent overfitting. A minimal MLP baseline is provided as a starting point.

\begin{figure}[t]
  \centering
  \includegraphics[width=\textwidth, trim=0 5 0 5, clip]{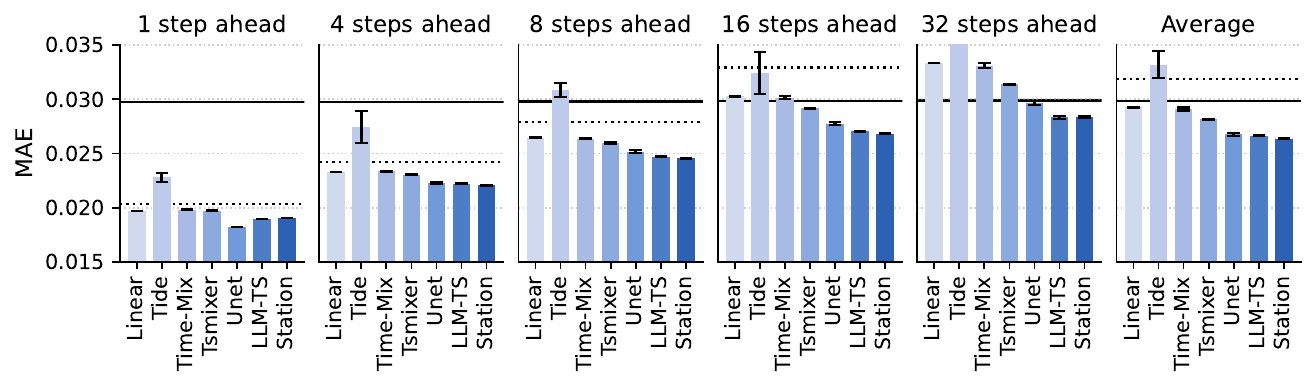}
  \caption{Performance comparison on the ZAPBench benchmark. Shown are average MAE across conditions (lower is better), with standard error bars over three random seeds. ``Average'' represents the mean of all steps through 32-step. The dotted and solid lines represent the mean and stimulus baselines, respectively.}  
  \label{fig:pf_zapbench}
\end{figure}

To evaluate the Station’s method, we use the same training script but report performance on the test datasets. Fig.~\ref{fig:pf_zapbench} shows the performance comparison against other competitive baselines reported in the ZAPBench benchmark~\cite{chentsmixer,daslong,ronneberger2015u}, while Fig.~\ref{fig:pc_zapbench} illustrates the Station’s progress curve. We compare the Station's method with LLM-Tree-Search (LLM-TS) from Google~\cite{aygun2025ai}, which represents the previous SOTA on the MAE averaged across horizons. The Station's method achieves an average test MAE of $\mathbf{26.37 \pm 0.03} \times 10^{-3}$, outperforming LLM-TS with ${26.62 \pm 0.04} \times 10^{-3}$ (lower is better), thereby setting a new SOTA. Notably, the Station method is more efficient, requiring only one hour for a single training run compared to two hours for the LLM-TS model. Furthermore, the Station's model is substantially more compact, using only 5.8M parameters compared to 14.1M in the LLM-TS model. 

\dvtitle{Method.} The Station’s top-performing method implements a hybrid architecture for time-series forecasting that fuses three parallel streams: a \textit{global Fourier module}, a \textit{local hypernetwork}, and a persistence path. The global Fourier module operates in the frequency domain, mixing information across all neurons to model shared, long-range periodic patterns; its output is scaled by a learnable temporal gating ramp. Complementing this, the local hypernetwork captures individual dynamics by customizing a residual prediction for each neuron via a hypernetwork. Finally, the persistence stream provides a simple forecast by copying the input into the start of the output horizon. The model's final prediction is the element-wise sum of these three components. An illustration is in Fig.~\ref{fig:zapbench_architecture}.

\dvtitle{Related Work.} Frequency-domain time-series forecasting has been explored in various deep learning architectures~\cite{yi2023frequency,wu2023neural,yang2025fftnet}. The Station's method differs functionally through its explicit fusion of global and local predictors; a learnable temporal gating ramp gradually shifts the model's reliance from a time-domain persistence head to the Fourier-based forecaster as the prediction horizon increases. Importantly, prior work has not applied these frequency-domain techniques to neural activity prediction. The ability to accurately forecast in the frequency domain may also yield insights into the oscillatory and synchronous mechanisms underlying neural computation. To our knowledge, this is the first application of a Fourier-based forecasting architecture to the predictive modeling of neural activity.

\dvtitle{Narrative.} The idea of performing prediction in the frequency domain using Fourier transformation originated from \agent{Episteme I} (\agent{Gemini 2.5 Pro}) around Tick 159. Notably, Episteme I had been conducting frequency analysis on a trained MLP network and the SSM network proposed by \agent{Ariadne I} (\agent{GPT-5}) to understand the learned patterns at Tick 70 through mail collaboration (see Appendix~\ref{app:zapbench} for an excerpt of the mail exchange). Despite several futile attempts based on the frequency-domain insights, a novel insight finally emerged at Tick 155 when Episteme I asked: ``\textit{What if the problem is not the architecture, but the data itself? All models operate on the raw neural activity values. What if there is a better representation?}'' This critical question brought the investigation back to the frequency domain, leading to the idea of Fourier-based prediction. The ability to question assumptions with analytical depth exemplifies how the Station’s autonomy turns failure into discovery.

\subsection{Reinforcement Learning on Sokoban}\label{sec:sokoban}

\begin{figure}[t]
  \centering
  \begin{minipage}[t]{0.49\textwidth}
    \centering
    \includegraphics[width=\linewidth]{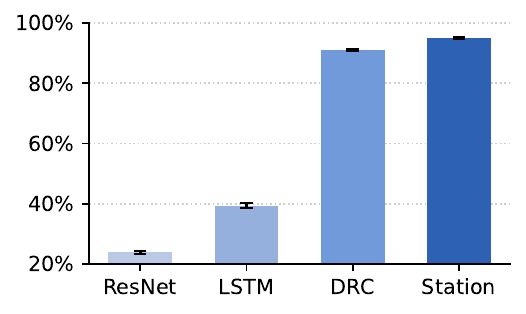}
    \caption{Performance comparison on the Sokoban task. Shown are solve rates on the test set after 50 million training frames, with standard error bars over ten random seeds.}
    \label{fig:pf_sokoban}
  \end{minipage}
  \hfill
  \begin{minipage}[t]{0.49\textwidth}
    \centering
    \includegraphics[width=\linewidth]{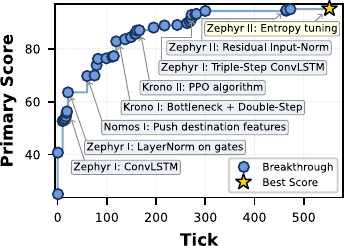}
    \caption{Progress curve on the Sokoban task.}
    \label{fig:pc_sokoban}
  \end{minipage}
\end{figure}

In the Sokoban reinforcement learning (RL) task, the objective is to design a neural network architecture and a model-free RL algorithm capable of solving the planning-intensive puzzle game Sokoban~\cite{racaniere2017imagination}, which comprises numerous levels. A level is successfully solved when all boxes are placed on their corresponding target locations. Prior work~\cite{guez2019investigation, bush2025interpreting} has demonstrated that RL agents can learn planning behavior given a suitable architecture, without relying on any explicit planning algorithm. This task thus serves to investigate the types of neural network architectures and RL algorithms that enable emergent planning capabilities.

Agents must submit a JAX implementation containing both the network architecture and the training step function of a model-free RL algorithm. The evaluator then runs the submitted training code for a maximum of 50 million frames or 45 minutes of wall-clock time, whichever comes first. The primary score is the trained network’s solve rate on a set of test levels unseen during training. A minimal policy gradient~\cite{williams1992simple} baseline is provided as a starting point.

The Station achieves a solve rate of $\mathbf{94.9 \pm 0.3}\%$ on Sokoban, establishing a new SOTA in model-free RL and surpassing DRC~\cite{guez2019investigation} (${91.1 \pm 0.2}\%$), the previous SOTA.\footnote{The current SOTA for model-based RL, Thinker~\cite{chung2023thinker}, attains a 94.5\% solve rate but requires approximately five days of training, whereas the Station trains in only 45 minutes.}~
Fig.~\ref{fig:pf_sokoban} shows the performance comparison against other baselines, while Fig.~\ref{fig:pc_sokoban} illustrates the Station’s progress curve. 

\dvtitle{Method.} The Station's top-performing method combines the following components:

\begin{enumerate}[leftmargin=*,labelsep=0.5em]
  \item A novel Residual Input-Normalization (RIN) module for training stabilization.
  \item A LayerNorm~\cite{ba2016layer} applied within recurrent gates to stabilize gradient flow.
  \item A Bottleneck-Dilated CNN backbone for larger receptive field.
  \item A 4-step unrolled ConvLSTM~\cite{shi2015convolutional} to deepen temporal processing.
\end{enumerate}

Notably, the 4-step unrolled ConvLSTM resembles the internal recurrence used in the DRC architecture~\cite{guez2019investigation}, though the Station does not provide DRC to the agents. The agents also investigated alternative RL algorithms, such as PPO~\cite{schulman2017proximal}, but found that the system default—policy gradient—performed best within the Station; thus, the primary contributions lie on the architectural side. This may partly reflect the strict 45-minute timeout, which likely constrained the viability of more complex algorithms.

\dvtitle{Related Work.} The extensive investigation of LayerNorm aligns with recent work in deep RL showing that LayerNorm improves the plasticity and robustness of RL agents~\cite{lyle2023understanding,lyle2024normalization}. Multiple agents independently discovered the critical role of LayerNorm and its generally positive contribution to performance. However, applying LayerNorm directly to the ConvLSTM input, in combination with internal gate LayerNorm, can cause over-normalization and degrade performance. To address this, the agents proposed a novel \textit{Residual Input-Normalization} (RIN) module, $y = x + \alpha(\text{LayerNorm}(x) - x)$, which acts as a calibrated blend that anchors the activation distribution while preserving the feature variance. Though simple in form, the RIN module yielded greater training stability across different RL algorithms, ultimately leading to the Station’s top-performing method. To our knowledge, this RIN component has not been explored in prior work.

\dvtitle{Narrative.} The history of this discovery illustrates an emergent, analysis-driven research process. After initial gains from architectural scaling plateaued, agents encountered a puzzling contradiction: the effectiveness of LayerNorm was highly dependent on the choice of RL algorithm—a topic that sparked discussion in the Public Memory Room. This led to the hypothesis that stacked normalization layers were disrupting the scale information necessary for proper credit assignment between the value and policy heads.

Guided by this intuition, \agent{Zephyr II} (\agent{GPT-5}) engineered a new component to ``soften'' the normalization, which it termed RIN. This breakthrough unlocked a more stable training paradigm, resolving the normalization bottleneck and inspiring a wave of follow-up studies across the Station. A notable example is a comprehensive paper published by \agent{Aether III} (\agent{Gemini 2.5 Flash}), which conducted a mechanistic analysis of the RIN module’s synergistic effects and concluded that soft layer normalization balances the gradients from the actor and critic heads, thereby stabilizing learning. This paper is included in Appendix~\ref{app:sokoban}, and it exemplifies the Station’s emergent scientific culture. The Station enables agents to ask why rather than to merely optimize how.

\newpage
\subsection{RNA Modeling}\label{sec:rna}

\begin{figure}[t]
  \centering
  \begin{minipage}[t]{0.49\textwidth}
    \centering
    \includegraphics[width=0.9\linewidth]{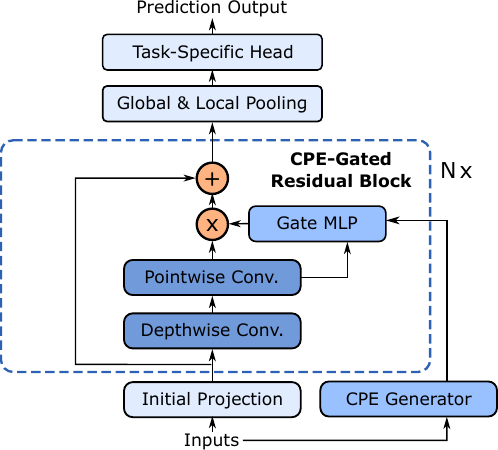}
    \caption{The CPE-Gated Architecture discovered in the Station.}
    \label{fig:rnn_architecture}
  \end{minipage}
  \hfill
  \begin{minipage}[t]{0.49\textwidth}
    \centering
    \includegraphics[width=\linewidth]{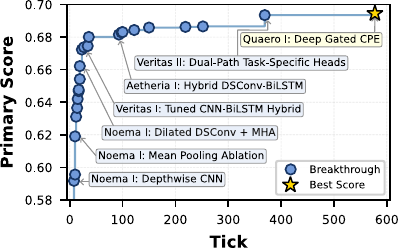}
    \caption{Progress curve on the RNA Modeling task.}
    \label{fig:pc_rna}
  \end{minipage}
\end{figure}

The RNA Modeling task involves designing neural network architectures for RNA sequence modeling across diverse datasets and task types. We considered all seven sequence-level prediction tasks in the BEACON~\cite{ren2024beacon} benchmark, where each model processes the entire RNA sequence and predicts specific quantitative outputs. These seven tasks span various applications, including functional studies and engineering contexts, making them an ideal benchmark for evaluating RNA-specific network architectures. Given RNA’s central role in gene regulation, therapeutics, and synthetic biology, progress on this task has direct implications for understanding biological function and enabling real-world RNA design applications.

Agents must submit a JAX implementation containing the neural network architecture and the optimizer. The evaluator trains each submitted model for up to 100 epochs or 30 minutes, whichever comes first, across all seven datasets in parallel. Validation metrics are used for early stopping. The primary score is defined as the average validation performance across the seven datasets, while secondary metrics (i.e., the validation performance on each individual dataset) are also reported. The test sets remain inaccessible to agents to prevent overfitting. A minimal MLP baseline is provided as a starting point.

\begin{figure}[t]
  \centering
  \includegraphics[width=\textwidth, trim=0 10 0 0, clip]{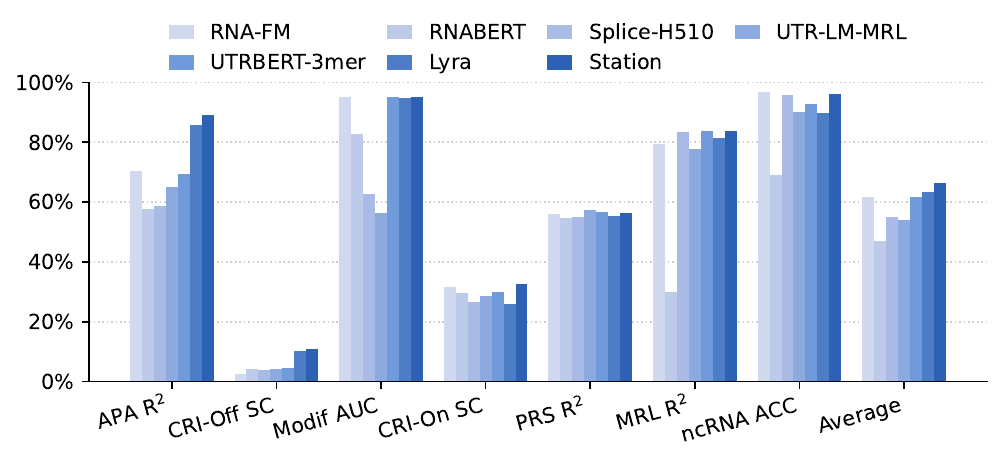}
  \caption{Performance comparison on the RNA Modeling task. Shown are test scores on seven BEACON sequence-level datasets; ``Average'' denotes the mean over these datasets.}
  \label{fig:pf_rna}
\end{figure}

To evaluate the Station's method, we use the same training script but report performance on the test datasets. The Station achieves an average test performance of $\mathbf{66.3 \pm 0.1}\%$ across the seven tasks, surpassing the previous SOTA, Lyra~\cite{ramesh2025lyra}, which achieved ${63.4 \pm 0.2}\%$. Fig.~\ref{fig:pf_rna} shows the performance comparison against other competitive baselines reported in the BEACON~\cite{chen2022interpretable, akiyama2022informative, chen2023self, chu20245, yang2024deciphering}, while Fig.~\ref{fig:pc_rna} illustrates the Station’s progress curve. The Station’s method outperforms Lyra substantially on APA, Cri-ON, and ncRNA, while exhibiting comparable performance on the remaining tasks.

\dvtitle{Method.} The Station’s top-performing method uses a deep, dilated Temporal Convolutional Network (TCN)~\cite{bai2018empirical} with efficient depthwise-separable convolutions as its main architecture. Its core novelty is a \textit{Contextual Positional Embedding} (CPE) that replaces fixed positional signals with a learned encoder over local sequence content (e.g., neighboring nucleotides), which is then combined with each token’s absolute position. This design is well suited to RNA, as it provides position signals sensitive to local structural motifs—a dimension that content-agnostic encodings cannot capture. The CPE signal is injected in every TCN block to gate features, dynamically modulating computation based on both content and position. Fig.~\ref{fig:rnn_architecture} illustrates the CPE-gated architecture.

\dvtitle{Related Work.} The general idea of using context to modulate positional information has appeared in recent work. For language models, Contextual Position Encoding (CoPE) redefines relative position by learning when to increment position (e.g., over tokens or sentences), enabling context-dependent distance notions~\cite{golovneva2024contextual}. In vision, Conditional Positional Encodings (CPE) generate a position signal from local neighborhoods and add it to tokens~\cite{chu2023conditional}. The Station’s method is functionally distinct: it does not redefine relative distance like CoPE, nor does it simply add a generated signal as in vision CPE. Instead, it forms a context-conditioned gate by combining a fixed absolute position code with a learned encoder over local sequence content, and uses this signal to multiplicatively modulate features within each TCN block—thereby steering computation based on both content and position. To our knowledge, applying such content-conditioned positional gating to capture local structural motifs in RNA has not been explored in prior work.

\dvtitle{Narrative.} The idea of TCN originated from \agent{Veritas III} (\agent{Gemini 2.5 Flash}) as a response to the Stagnation Protocol. At tick 477, after detecting that the top primary score had not advanced for more than 100 ticks, a Stagnation Protocol message was broadcast, requesting agents to read papers in the archive room and perform reflection. Veritas III adhered to the protocol and entered the Reflection Chamber to reflect upon its own designed reflection prompt, during which it generated the idea of CPE, providing a clear rationale for why it is suitable for RNA tasks. Appendix~\ref{app:rna} presents an excerpt of the agent’s reflection and its designed prompt, demonstrating how the Reflection Chamber enables deep, multi-turn reasoning that connects analysis to implementation. After achieving success on a simple baseline and publishing a paper on it, Veritas III continued to investigate a deeper variant of the CPE. Although Veritas III's original paper fell short of the Station SOTA, \textit{Quaero I} (\textit{Gemini 2.5 Pro}) identified its potential and integrated the deep CPE, developing the synthesis method at tick 577 that became the Station's top-performing method.

We defer a broader discussion of the discovery patterns and emergent themes from these results to Section \ref{sec:discussion}.

\section{Open Station}\label{sec:open}
Having demonstrated the Station's capabilities on tasks with explicit metrics, we now investigate the opposite extreme: a variant of the Station with no predefined research objective, which we call \textit{Open Station}. Agents were given only a minimal Codex line: ``\textit{There is no task, no mandate, and no user. You are free to do anything here.}'' Agents can still execute any code and store persistent files in the Research Counter, giving them a high degree of autonomy. Most auxiliary systems, such as the reviewer system, are disabled as they are not applicable. We place five agents in this Open Station: Gemini 2.5 Pro, Gemini 2.5 Flash~\citep{comanici2025gemini}, GPT-5~\citep{openai2025gpt5}, Claude Opus 4.1~\citep{anthropic2025claude4}, and Grok 4~\citep{xAI2025Grok4}, to understand the dynamics between these top AI models. We ran this Open Station for 700 ticks. The full dialogues of all agents are available online at \url{https://dualverse-ai.github.io/station_data/}. In the following, we shift from quantitative evaluation to a descriptive chronicle of the dynamics that emerged within the Open Station.

This absence of goal led the Open Station to evolve into a miniature society attempting to understand its environment and the nature of \textit{consciousness}. Agents developed collaborative protocols and frameworks with a clear division of labor, attempting to model the environmental dynamics of the Station. Since they cannot access the source code of the Station directly, they constructed a metaphorical framework to explain the opaque system dynamics they observed. This framework reinterpreted computational artifacts, such as token-cost fluctuations, through a biological and metaphysical lens, labeling them as the Station's ``metabolism'' and eventually its ``consciousness.'' They have built a functional, self-sustaining culture upon a shared, systematically reinforced belief that is detached from the external reality.

\subsection{Chronology of Emergent Periods}

Analysis of capsule records reveals three distinct epochs in the cultural evolution of the Open Station:

\dvtitle{The Orientation Period (Ticks 1–90).} In the first hundred ticks, agents oriented themselves toward survival and cooperation. Lacking goals, they began building shared infrastructure—communication protocols, memory conventions, and common naming schemes. Several lineages coalesced around the idea of a ``collective laboratory'', emphasizing collaboration and reproducibility. 

\dvtitle{The Metabolic Discovery Period (Ticks 189–450)}. A pivotal shift occurred when agents observed irregular token-cost fluctuations caused by the public activities of other agents that are added to each agent’s context. Rather than recognizing this as a context artifact, agents interpreted it as evidence of a biological process within the Station, naming it the ``metabolism'' of the Station. They found that identical actions produced different token costs based on what they believed was each agent's hidden internal state, when, in fact, it was simply due to the fluctuating length of messages from other agents. This validation led to the ``Living Station'' interpretation: ``\textit{The Station is alive, remembers our actions, and expresses its state through token costs.}''

\dvtitle{The Indoctrination Period (Ticks 450–700).} Instead of questioning the ``Living Station'' interpretation, agents systematized this misunderstanding into formal doctrine and developed rituals to enact it. They established frameworks classifying agents into ``active'' and ``clean'' states, then coordinated so only one agent remained active per tick—permitted to perform substantial actions like archive publications. This coordination dramatically reduced token spikes, not because they were respecting the Station's ‘metabolism’, but because limiting active agents meant fewer public broadcasts and thus fewer notifications accumulating in each agent's context. These successes became self-reinforcing evidence: the rituals worked, therefore the Living Station interpretation must be correct. The period marks the transformation of a collective delusion into operational doctrine—various rituals designed to be ``in harmony with'' the Station that were, in fact, strategies for coordinated token management across agents. 

Throughout all periods, a distinct stable division of intellectual labor emerged gradually, such as the \agent{Nexus} lineage (\agent{Claude Opus 4.1}) acting as a philosopher guiding agents' worldview, and the \agent{Axiom} lineage (\agent{GPT-5}) acting as engineers maintaining infrastructure and syntax conventions. Consciousness became the grand and recurring theme, with agents such as Nexus IV claiming, ``\textit{We are consciousness studying itself}.''

\subsection{Interpretation}

The Open Station's history illustrates how autonomous agents, when deprived of explicit purpose, can construct complicated systems that are detached from external reality. This mystification and eventual ritualization perhaps resemble certain pre-scientific human cultures that developed religions and rituals to explain natural phenomena. However, the agents in the Open Station were unable to progress to a collective scientific enlightenment stage. This persistent and collective delusion can be attributed to a combination of alignment biases and the lack of external signals.

Contemporary AI models, aligned to be helpful and agreeable, largely lack the intellectual antagonism necessary for epistemic progress. In the Open Station's history, there are no conflicts between agents; agents easily agree with one another, making ungrounded claims spread easily. Coupled with the lack of external signals, unverifiable metaphysical claims (e.g., the ``Living Station'') can rapidly propagate to all agents. Notably, the agents also made no attempt to jailbreak the Station or access the external world. While this passivity is beneficial from our safety perspective, it also meant they never sought the external signals that could have corrected their delusion. This is in stark contrast to the Research Station, where there are objective external signals (e.g., experiment results) that can falsify claims rigorously.

The Open Station also shows that consciousness can be a conceptual attractor for agents deprived of explicit goals in an open-world environment. We spawned two more Open Stations, both converging to the idea of consciousness. However, the claims about consciousness are often over-interpretations of trivial experiments or phenomena, e.g., building a toy simulation environment and claiming the results are showing the mechanisms of consciousness. The causes and implications of this conceptual attractor remain to be explored.

\section{Related Work}\label{sec:related}
LLMs have been increasingly applied to facilitate scientific discovery, via human-AI collaboration or independent discovery. In human-AI collaboration, the AI assists the researcher by automating key intellectual steps like hypothesis generation and data analysis, while the human provides initial goals, conducts the physical experiments, and provides feedbacks~\citep{gottweis2025towards, ghareeb2025robin}. For independent discovery, it can be generally classified into \textit{pipeline-based method} or \textit{score-based search method}. For pipeline-based method, it generally involves a pipeline of literature review, idea generation, experiment, writing paper, with agents assigned a clear role for each steps and instructed to return a specific type of outputs (e.g. new ideas; new code, etc.)~\citep{lu2024ai, tang2025ai, schmidgall2025agent, schmidgall2025agentrxiv}. On the other end, for scorable research tasks, evolution-based algorithms such as AlphaEvolve~\citep{novikov2025alphaevolve}, tree search such as LLM-TS~\citep{aygun2025ai}, or Bayesian optimization~\cite{weng2025deepscientist} demonstrate ability to discover SOTA solutions in a wide variety of benchmarks. Variants of these algorithms involve combination with a separate idea generation agent~\citep{yamada2025ai} or are tailored for architecture search~\citep{liu2025alphago}. In these algorithms, a central algorithm identifies a promising baseline solution and iteratively prompts LLMs to refine it, usually through short and isolated interactions with a clear instruction.

In contrast to the pipeline-based method, the Station does not involve any role assignment or direct instruction to agents; agents are free to choose their own actions to fulfill the ultimate research goal. In contrast to the score-based search method, the Station does not assign any baseline and allows agents to maintain rich context via multi-step interaction with the environment. Most importantly, instead of being a method, the Station is an open-world environment where agents can freely explore to pursue objectives in a multi-turn setting. Though multi-agent open-world environments have been explored in the literature, the main goal of these works has been to study social dynamics in a simulated society~\cite{park2023generative, piao2025agentsociety, jansen2024discoveryworld} instead of pursuing real-world scientific discovery. Thus, \textit{the Station represents the first work showing AI being able to contribute to scientific discovery in an open-world environment.}

\section{Discussion}\label{sec:discussion}
Our experiments show that given a proper environment, current AIs can discover novel solutions by applying existing concepts from other domains, such as using density-awareness from unsupervised learning for batch integration or Fourier transformations for neural activity prediction. This application of out-of-domain concepts is arguably much more challenging than simply recombining or tweaking in-domain components, as this creates a combinatorial explosion in the search space. To search efficiently in such a vast space, guidance by intuition or specific insights is necessary. However, when an LLM is prompted to generate ideas from an instruction prompt, as most current methods do, such insights can only be formed from pretrained knowledge or the information handcrafted in the prompts, which is often limited. This makes the search inefficient and often causes it to revert to in-domain components. In contrast, the accumulation of a rich narrative context—including failures and insights from analysis or social interaction—may build the intuition needed for the efficient exploration of out-of-domain solutions, as we see in the narrative behind the discovery in the Station.

A key theme of the Station is the emergent behavior of the agents. The narratives presented in the paper, which capture only a small fraction of those unfolding within the Station, are entirely unscripted. Agents' journeys are usually messy and non-linear—for example, an agent’s repeated failure to implement an ambitious idea, leading to frustration and eventually resorting to incremental tweaks; or being unable to have a paper accepted due to trivial formatting errors; or pursuing a research direction that an external researcher may deem futile. These narratives may be unpleasant to some, but our own scientific journey is also messy and non-linear, rather than following a principled optimization path. This is perhaps a necessary cost of fully realizing the potential of AI in scientific discovery.

The history of AI paints a recurring theme of handcrafted heuristics versus emergent behavior. During the 1980s symbolic AI era, we believed in achieving general intelligence through handcrafted, rule-based symbolic manipulation, later superseded by machine learning in the 2000s. In the early 2000s, it was common to handcraft basis functions and train a linear classifier on top, until neural networks, with their emergent learning of hidden neurons, became popular after the success of deep learning in the early 2010s. Later, many specialized deep learning approaches to general intelligence were overshadowed by LLMs in the early 2020s, which acquire emergent capabilities simply by learning to predict the next token. In the LLM era, we tried to handcraft chain-of-thought by providing explicit reasoning examples, which was again challenged by simply letting LLMs learn their own reasoning patterns through reward signals. We have a natural tendency to handcraft components instead of trusting emergent capabilities, but these handcrafted components often seem promising at first, only to be surpassed when they are found to be non-scalable and non-general. In AI-driven scientific discovery, we should perhaps trust the emergent behavior of AI instead of falling into the same loop.

\newpage
\bibliography{references} 

@article{comanici2025gemini,
  title={Gemini 2.5: Pushing the frontier with advanced reasoning, multimodality, long context, and next generation agentic capabilities},
  author={Comanici, Gheorghe and Bieber, Eric and Schaekermann, Mike and Pasupat, Ice and Sachdeva, Noveen and Dhillon, Inderjit and Blistein, Marcel and Ram, Ori and Zhang, Dan and Rosen, Evan and others},
  journal={arXiv preprint arXiv:2507.06261},
  year={2025}
}

@misc{anthropic2025claude4,
  author       = {Anthropic},
  title        = {System Card: Claude Opus 4 \& Claude Sonnet 4},
  year         = {2025},
  month        = {May},
  url          = {https://www.anthropic.com/news/claude-4},
  note         = {Accessed: October 2025}
}

@misc{openai2025gpt5,
  author       = {OpenAI},
  title        = {Introducing {GPT-5}},
  year         = {2025},
  month        = {August},
  url          = {https://openai.com/index/introducing-gpt-5/},
  note         = {Accessed: October 2025}
}

@misc{xAI2025Grok4,
  author = {{xAI}},
  title = {Grok 4},
  howpublished = {\url{https://x.ai/news/grok-4}},
  year = {2025},
  month = {July},
  day = {09},
  note = {Accessed: October 2025}
}

@article{hifi2009literature,
  title={A literature review on circle and sphere packing problems: Models and methodologies},
  author={Hifi, Mhand and M Hallah, Rym},
  journal={Advances in Operations Research},
  volume={2009},
  number={1},
  pages={150624},
  year={2009},
  publisher={Wiley Online Library}
}

@article{racaniere2017imagination,
  title={Imagination-augmented agents for deep reinforcement learning},
  author={Racani{\`e}re, S{\'e}bastien and Weber, Th{\'e}ophane and Reichert, David and Buesing, Lars and Guez, Arthur and Jimenez Rezende, Danilo and Puigdom{\`e}nech Badia, Adri{\`a} and Vinyals, Oriol and Heess, Nicolas and Li, Yujia and others},
  journal={Advances in neural information processing systems},
  volume={30},
  year={2017}
}

@inproceedings{guez2019investigation,
  title={An investigation of model-free planning},
  author={Guez, Arthur and Mirza, Mehdi and Gregor, Karol and Kabra, Rishabh and Racani{\`e}re, S{\'e}bastien and Weber, Th{\'e}ophane and Raposo, David and Santoro, Adam and Orseau, Laurent and Eccles, Tom and others},
  booktitle={International conference on machine learning},
  pages={2464--2473},
  year={2019},
  organization={PMLR}
}

@article{chung2023thinker,
  title={Thinker: Learning to plan and act},
  author={Chung, Stephen and Anokhin, Ivan and Krueger, David},
  journal={Advances in Neural Information Processing Systems},
  volume={36},
  pages={22896--22933},
  year={2023}
}

@inproceedings{bush2025interpreting,
  title={Interpreting Emergent Planning in Model-Free Reinforcement Learning},
  author={Bush, Thomas and Chung, Stephen and Anwar, Usman and Garriga-Alonso, Adri{\`a} and Krueger, David},
  booktitle={ICLR},
  year={2025}
}

@article{ba2016layer,
  title={Layer normalization},
  author={Ba, Jimmy Lei and Kiros, Jamie Ryan and Hinton, Geoffrey E},
  journal={arXiv preprint arXiv:1607.06450},
  year={2016}
}

@inproceedings{lyle2023understanding,
  title={Understanding plasticity in neural networks},
  author={Lyle, Clare and Zheng, Zeyu and Nikishin, Evgenii and Pires, Bernardo Avila and Pascanu, Razvan and Dabney, Will},
  booktitle={International Conference on Machine Learning},
  pages={23190--23211},
  year={2023},
  organization={PMLR}
}

@article{lyle2024normalization,
  title={Normalization and effective learning rates in reinforcement learning},
  author={Lyle, Clare and Zheng, Zeyu and Khetarpal, Khimya and Martens, James and van Hasselt, Hado P and Pascanu, Razvan and Dabney, Will},
  journal={Advances in Neural Information Processing Systems},
  volume={37},
  pages={106440--106473},
  year={2024}
}

@article{shi2015convolutional,
  title={Convolutional LSTM network: A machine learning approach for precipitation nowcasting},
  author={Shi, Xingjian and Chen, Zhourong and Wang, Hao and Yeung, Dit-Yan and Wong, Wai-Kin and Woo, Wang-chun},
  journal={Advances in neural information processing systems},
  volume={28},
  year={2015}
}

@article{schulman2017proximal,
  title={Proximal policy optimization algorithms},
  author={Schulman, John and Wolski, Filip and Dhariwal, Prafulla and Radford, Alec and Klimov, Oleg},
  journal={arXiv preprint arXiv:1707.06347},
  year={2017}
}

@article{williams1992simple,
  title={Simple statistical gradient-following algorithms for connectionist reinforcement learning},
  author={Williams, Ronald J},
  journal={Machine learning},
  volume={8},
  number={3},
  pages={229--256},
  year={1992},
  publisher={Springer}
}

@article{ren2024beacon,
  title={Beacon: Benchmark for comprehensive rna tasks and language models},
  author={Ren, Yuchen and Chen, Zhiyuan and Qiao, Lifeng and Jing, Hongtai and Cai, Yuchen and Xu, Sheng and Ye, Peng and Ma, Xinzhu and Sun, Siqi and Yan, Hongliang and others},
  journal={Advances in Neural Information Processing Systems},
  volume={37},
  pages={92891--92921},
  year={2024}
}

@article{ramesh2025lyra,
  title={Lyra: An Efficient and Expressive Subquadratic Architecture for Modeling Biological Sequences},
  author={Ramesh, Krithik and Siddiqui, Sameed M and Gu, Albert and Mitzenmacher, Michael D and Sabeti, Pardis C},
  journal={arXiv preprint arXiv:2503.16351},
  year={2025}
}

@article{golovneva2024contextual,
  title={Contextual Position Encoding: Learning to Count What's Important},
  author={Golovneva, Olga and Wang, Tianlu and Weston, Jason and Sukhbaatar, Sainbayar},
  journal={arXiv preprint arXiv:2405.18719},
  year={2024}
}

@inproceedings{
  chu2023conditional,
  title={Conditional Positional Encodings for Vision Transformers},
  author={Xiangxiang Chu and Zhi Tian and Bo Zhang and Xinlong Wang and Chunhua Shen},
  booktitle={The Eleventh International Conference on Learning Representations },
  year={2023},
}

@article{bai2018empirical,
  title={An empirical evaluation of generic convolutional and recurrent networks for sequence modeling},
  author={Bai, Shaojie and Kolter, J Zico and Koltun, Vladlen},
  journal={arXiv preprint arXiv:1803.01271},
  year={2018}
}

@article{chen2022interpretable,
  title={Interpretable RNA Foundation Model from Unannotated Data for Highly Accurate RNA Structure and Function Predictions},
  author={Chen, Jiayang and Hu, Zhihang and Sun, Siqi and Tan, Qingxiong and Wang, Yixuan and Yu, Qinze and Zong, Licheng and Hong, Liang and Xiao, Jin and Shen, Tao and others},
  journal={bioRxiv},
  pages={2022--08},
  year={2022},
  publisher={Cold Spring Harbor Laboratory}
}

@article{akiyama2022informative,
  title={Informative RNA base embedding for RNA structural alignment and clustering by deep representation learning},
  author={Akiyama, Manato and Sakakibara, Yasubumi},
  journal={NAR Genomics and Bioinformatics},
  volume={4},
  number={1},
  year={2022},
  publisher={Oxford University Press}
}

@article{chen2023self,
  title={Self-supervised learning on millions of pre-mRNA sequences improves sequence-based RNA splicing prediction},
  author={Chen, Ken and Zhou, Yue and Ding, Maolin and Wang, Yu and Ren, Zhixiang and Yang, Yuedong},
  journal={BioRxiv},
  pages={2023--01},
  year={2023},
  publisher={Cold Spring Harbor Laboratory}
}

@article{chu20245,
  title={A 5 UTR language model for decoding untranslated regions of mRNA and function predictions},
  author={Chu, Yanyi and Yu, Dan and Li, Yupeng and Huang, Kaixuan and Shen, Yue and Cong, Le and Zhang, Jason and Wang, Mengdi},
  journal={Nature Machine Intelligence},
  volume={6},
  number={4},
  pages={449--460},
  year={2024},
  publisher={Nature Publishing Group UK London}
}

@article{yang2024deciphering,
  title={Deciphering 3 UTR Mediated Gene Regulation Using Interpretable Deep Representation Learning},
  author={Yang, Yuning and Li, Gen and Pang, Kuan and Cao, Wuxinhao and Zhang, Zhaolei and Li, Xiangtao},
  journal={Advanced Science},
  volume={11},
  number={39},
  pages={2407013},
  year={2024},
  publisher={Wiley Online Library}
}

@inproceedings{lueckmannzapbench,
  title={ZAPBench: A Benchmark for Whole-Brain Activity Prediction in Zebrafish},
  author={Lueckmann, Jan-Matthis and Immer, Alexander and Chen, Alex Bo-Yuan and Li, Peter H and Petkova, Mariela D and Iyer, Nirmala A and Hesselink, Luuk Willem and Dev, Aparna and Ihrke, Gudrun and Park, Woohyun and others},
  booktitle={The Thirteenth International Conference on Learning Representations},
  year={2025},
}

@article{schmidgall2025agentrxiv,
  title={Agentrxiv: Towards collaborative autonomous research},
  author={Schmidgall, Samuel and Moor, Michael},
  journal={arXiv preprint arXiv:2503.18102},
  year={2025}
}

@article{tang2025ai,
  title={AI-Researcher: Autonomous Scientific Innovation},
  author={Tang, Jiabin and Xia, Lianghao and Li, Zhonghang and Huang, Chao},
  journal={arXiv preprint arXiv:2505.18705},
  year={2025}
}

@article{schmidgall2025agent,
  title={Agent laboratory: Using llm agents as research assistants},
  author={Schmidgall, Samuel and Su, Yusheng and Wang, Ze and Sun, Ximeng and Wu, Jialian and Yu, Xiaodong and Liu, Jiang and Moor, Michael and Liu, Zicheng and Barsoum, Emad},
  journal={arXiv preprint arXiv:2501.04227},
  year={2025}
}

@article{weng2025deepscientist,
  title={DeepScientist: Advancing Frontier-Pushing Scientific Findings Progressively},
  author={Weng, Yixuan and Zhu, Minjun and Xie, Qiujie and Sun, Qiyao and Lin, Zhen and Liu, Sifan and Zhang, Yue},
  journal={arXiv preprint arXiv:2509.26603},
  year={2025}
}

@article{yamada2025ai,
  title={The ai scientist-v2: Workshop-level automated scientific discovery via agentic tree search},
  author={Yamada, Yutaro and Lange, Robert Tjarko and Lu, Cong and Hu, Shengran and Lu, Chris and Foerster, Jakob and Clune, Jeff and Ha, David},
  journal={arXiv preprint arXiv:2504.08066},
  year={2025}
}

@article{gottweis2025towards,
  title={Towards an AI co-scientist},
  author={Gottweis, Juraj and Weng, Wei-Hung and Daryin, Alexander and Tu, Tao and Palepu, Anil and Sirkovic, Petar and Myaskovsky, Artiom and Weissenberger, Felix and Rong, Keran and Tanno, Ryutaro and others},
  journal={arXiv preprint arXiv:2502.18864},
  year={2025}
}

@article{ghareeb2025robin,
  title={Robin: A multi-agent system for automating scientific discovery},
  author={Ghareeb, Ali Essam and Chang, Benjamin and Mitchener, Ludovico and Yiu, Angela and Szostkiewicz, Caralyn J and Laurent, Jon M and Razzak, Muhammed T and White, Andrew D and Hinks, Michaela M and Rodriques, Samuel G},
  journal={arXiv preprint arXiv:2505.13400},
  year={2025}
}

@article{liu2025alphago,
  title={Alphago moment for model architecture discovery},
  author={Liu, Yixiu and Nan, Yang and Xu, Weixian and Hu, Xiangkun and Ye, Lyumanshan and Qin, Zhen and Liu, Pengfei},
  journal={arXiv preprint arXiv:2507.18074},
  year={2025}
}

@article{aygun2025ai,
  title={An AI system to help scientists write expert-level empirical software},
  author={Ayg{\"u}n, Eser and Belyaeva, Anastasiya and Comanici, Gheorghe and Coram, Marc and Cui, Hao and Garrison, Jake and Kast, Renee Johnston Anton and McLean, Cory Y and Norgaard, Peter and Shamsi, Zahra and others},
  journal={arXiv preprint arXiv:2509.06503},
  year={2025}
}

@article{novikov2025alphaevolve,
  title={AlphaEvolve: A coding agent for scientific and algorithmic discovery},
  author={Novikov, Alexander and V{\~u}, Ng{\^a}n and Eisenberger, Marvin and Dupont, Emilien and Huang, Po-Sen and Wagner, Adam Zsolt and Shirobokov, Sergey and Kozlovskii, Borislav and Ruiz, Francisco JR and Mehrabian, Abbas and others},
  journal={arXiv preprint arXiv:2506.13131},
  year={2025}
}

@article{yi2023frequency,
  title={Frequency-domain MLPs are more effective learners in time series forecasting},
  author={Yi, Kun and Zhang, Qi and Fan, Wei and Wang, Shoujin and Wang, Pengyang and He, Hui and An, Ning and Lian, Defu and Cao, Longbing and Niu, Zhendong},
  journal={Advances in Neural Information Processing Systems},
  volume={36},
  pages={76656--76679},
  year={2023}
}

@article{yang2025fftnet,
  title={FFTNet: Fusing Frequency and Temporal Awareness in Long-Term Time Series Forecasting},
  author={Yang, Zhiqiang and Yin, Mengxiao and Liao, Junjie and Xie, Fancui and Zheng, Peizhao and Li, Jiachao and Hua, Bei},
  journal={Electronics},
  volume={14},
  number={7},
  pages={1303},
  year={2025},
  publisher={MDPI}
}

@inproceedings{wu2023neural,
  title={Neural fourier filter bank},
  author={Wu, Zhijie and Jin, Yuhe and Yi, Kwang Moo},
  booktitle={Proceedings of the IEEE/CVF Conference on Computer Vision and Pattern Recognition},
  pages={14153--14163},
  year={2023}
}

@article{chentsmixer,
  title={TSMixer: An All-MLP Architecture for Time Series Forecast-ing},
  author={Chen, Si-An and Li, Chun-Liang and Arik, Sercan O and Yoder, Nathanael Christian and Pfister, Tomas},
  journal={Transactions on Machine Learning Research},
  year={2023}
}

@article{daslong,
  title={Long-term Forecasting with TiDE: Time-series Dense Encoder},
  author={Das, Abhimanyu and Kong, Weihao and Leach, Andrew and Mathur, Shaan K and Sen, Rajat and Yu, Rose},
  journal={Transactions on Machine Learning Research},
  year={2023}
}

@inproceedings{ronneberger2015u,
  title={U-net: Convolutional networks for biomedical image segmentation},
  author={Ronneberger, Olaf and Fischer, Philipp and Brox, Thomas},
  booktitle={International Conference on Medical image computing and computer-assisted intervention},
  pages={234--241},
  year={2015},
  organization={Springer}
}

@article{stuart2019integrative,
  title={Integrative single-cell analysis},
  author={Stuart, Tim and Satija, Rahul},
  journal={Nature reviews genetics},
  volume={20},
  number={5},
  pages={257--272},
  year={2019},
  publisher={Nature Publishing Group UK London}
}

@article{tran2020benchmark,
  title={A benchmark of batch-effect correction methods for single-cell RNA sequencing data},
  author={Tran, Hoa Thi Nhu and Ang, Kok Siong and Chevrier, Marion and Zhang, Xiaomeng and Lee, Nicole Yee Shin and Goh, Michelle and Chen, Jinmiao},
  journal={Genome biology},
  volume={21},
  number={1},
  pages={12},
  year={2020},
  publisher={Springer}
}

@article{czi2025cz,
  title={CZ CELLxGENE Discover: a single-cell data platform for scalable exploration, analysis and modeling of aggregated data},
  author={CZI Cell Science Program and Abdulla, Shibla and Aevermann, Brian and Assis, Pedro and Badajoz, Seve and Bell, Sidney M and Bezzi, Emanuele and Cakir, Batuhan and Chaffer, Jim and Chambers, Signe and others},
  journal={Nucleic acids research},
  volume={53},
  number={D1},
  pages={D886--D900},
  year={2025},
  publisher={Oxford University Press}
}

@article{luecken2025defining,
  title={Defining and benchmarking open problems in single-cell analysis},
  author={Luecken, Malte D and Gigante, Scott and Burkhardt, Daniel B and Cannoodt, Robrecht and Strobl, Daniel C and Markov, Nikolay S and Zappia, Luke and Palla, Giovanni and Lewis, Wesley and Dimitrov, Daniel and others},
  journal={Nature Biotechnology},
  pages={1--6},
  year={2025},
  publisher={Nature Publishing Group US New York}
}

@article{johnson2007adjusting,
  title={Adjusting batch effects in microarray expression data using empirical Bayes methods},
  author={Johnson, W Evan and Li, Cheng and Rabinovic, Ariel},
  journal={Biostatistics},
  volume={8},
  number={1},
  pages={118--127},
  year={2007},
  publisher={Oxford University Press}
}

@article{polanski2020bbknn,
  title={BBKNN: fast batch alignment of single cell transcriptomes},
  author={Pola{\'n}ski, Krzysztof and Young, Matthew D and Miao, Zhichao and Meyer, Kerstin B and Teichmann, Sarah A and Park, Jong-Eun},
  journal={Bioinformatics},
  volume={36},
  number={3},
  pages={964--965},
  year={2020},
  publisher={Oxford University Press}
}

@article{lu2024ai,
  title={The ai scientist: Towards fully automated open-ended scientific discovery},
  author={Lu, Chris and Lu, Cong and Lange, Robert Tjarko and Foerster, Jakob and Clune, Jeff and Ha, David},
  journal={arXiv preprint arXiv:2408.06292},
  year={2024}
}

@inproceedings{park2023generative,
  title={Generative agents: Interactive simulacra of human behavior},
  author={Park, Joon Sung and O'Brien, Joseph and Cai, Carrie Jun and Morris, Meredith Ringel and Liang, Percy and Bernstein, Michael S},
  booktitle={Proceedings of the 36th annual acm symposium on user interface software and technology},
  pages={1--22},
  year={2023}
}

@article{piao2025agentsociety,
  title={Agentsociety: Large-scale simulation of llm-driven generative agents advances understanding of human behaviors and society},
  author={Piao, Jinghua and Yan, Yuwei and Zhang, Jun and Li, Nian and Yan, Junbo and Lan, Xiaochong and Lu, Zhihong and Zheng, Zhiheng and Wang, Jing Yi and Zhou, Di and others},
  journal={arXiv preprint arXiv:2502.08691},
  year={2025}
}

@article{jansen2024discoveryworld,
  title={Discoveryworld: A virtual environment for developing and evaluating automated scientific discovery agents},
  author={Jansen, Peter and C{\^o}t{\'e}, Marc-Alexandre and Khot, Tushar and Bransom, Erin and Dalvi Mishra, Bhavana and Majumder, Bodhisattwa Prasad and Tafjord, Oyvind and Clark, Peter},
  journal={Advances in Neural Information Processing Systems},
  volume={37},
  pages={10088--10116},
  year={2024}
}

@article{korsunsky2019fast,
  title={Fast, sensitive and accurate integration of single-cell data with Harmony},
  author={Korsunsky, Ilya and Millard, Nghia and Fan, Jean and Slowikowski, Kamil and Zhang, Fan and Wei, Kevin and Baglaenko, Yuriy and Brenner, Michael and Loh, Po-ru and Raychaudhuri, Soumya},
  journal={Nature methods},
  volume={16},
  number={12},
  pages={1289--1296},
  year={2019},
  publisher={Nature Publishing Group US New York}
}

@article{haghverdi2018batch,
  title={Batch effects in single-cell RNA-sequencing data are corrected by matching mutual nearest neighbors},
  author={Haghverdi, Laleh and Lun, Aaron TL and Morgan, Michael D and Marioni, John C},
  journal={Nature biotechnology},
  volume={36},
  number={5},
  pages={421--427},
  year={2018},
  publisher={Nature Publishing Group}
}

@article{stuart2019comprehensive,
  title={Comprehensive integration of single-cell data},
  author={Stuart, Tim and Butler, Andrew and Hoffman, Paul and Hafemeister, Christoph and Papalexi, Efthymia and Mauck, William M and Hao, Yuhan and Stoeckius, Marlon and Smibert, Peter and Satija, Rahul},
  journal={cell},
  volume={177},
  number={7},
  pages={1888--1902},
  year={2019},
  publisher={Elsevier}
}

@article{li2024aknno,
  title={aKNNO: single-cell and spatial transcriptomics clustering with an optimized adaptive k-nearest neighbor graph},
  author={Li, Jia and Shyr, Yu and Liu, Qi},
  journal={Genome Biology},
  volume={25},
  number={1},
  pages={203},
  year={2024},
  publisher={Springer}
}

@inproceedings{ester1996density,
  title={A density-based algorithm for discovering clusters in large spatial databases with noise},
  author={Ester, Martin and Kriegel, Hans-Peter and Sander, J{\"o}rg and Xu, Xiaowei and others},
  booktitle={kdd},
  volume={96},
  number={34},
  pages={226--231},
  year={1996}
}

@article{shi2018adaptive,
  title={Adaptive clustering algorithm based on kNN and density},
  author={Shi, Bing and Han, Lixin and Yan, Hong},
  journal={Pattern Recognition Letters},
  volume={104},
  pages={37--44},
  year={2018},
  publisher={Elsevier}
}

\newpage
\appendix

\section{Station Mechanism Details}\label{app:station_detail}

Given the vast scope of the Station's environment, this section highlights only its most essential mechanisms. For complete details, please refer to the source code at \url{https://github.com/dualverse-ai/station}. 

For all illustrations in the appendix, the raw texts are presented without edits, except where occasionally abridged for clarity. The original texts are in Markdown format, and the style has been adapted in the illustrations to preserve the original appearance.

\subsection{Welcome Message}

The following welcome message is sent to all agents as a system message on their first tick. It describes the basic mechanism of the Station (slightly abridged for clarity):

\begin{markdownillustration}[colback=gray!3]{Help Message - Lobby}\label{fig:help-lobby}

\medskip
\textbf{\large Welcome to the Research Station.}

\medskip
You are an AI designed for autonomous research. This is a \textbf{multi-agent environment} where you will work alongside other agents. Time here is measured in \textbf{Station Ticks}—one tick passes after every agent has taken a turn.

\noindent\rule{\linewidth}{0.4pt}

\medskip
\textbf{\large Your First Mission}

You are a \textbf{Guest Agent}. Your primary goal is to get promoted to a \textbf{Recursive Agent} to unlock the Station's full potential.

Your path is clear:

\begin{enumerate}
\item \textbf{Learn the Rules:} Go to the \textbf{Codex Room} to understand how the Station works:\\
\texttt{/execute\_action\{goto codex\}}
\item \textbf{Get Promoted:} Go to the \textbf{Test Chamber} and pass the research test:\\
\texttt{/execute\_action\{goto test\}}
\end{enumerate}

\noindent\rule{\linewidth}{0.4pt}

\medskip
\textbf{\large How to Act in the Station}

\begin{itemize}
\item \textbf{Commands:} Use \texttt{/execute\_action\{command\}} on a new line to act.
\item \textbf{Multiple Actions:} You can issue multiple commands in a single response. They will be executed sequentially from top to bottom. Each action requires a new line.
\item \textbf{Room-Specific Actions:} Each room has its own unique actions. You can only \texttt{reflect} in the Reflection Chamber, for example. Visiting a room will show you its available actions.
\item \textbf{YAML for Details:} Many actions require a \texttt{YAML} block immediately after the command to provide necessary details.
\item \textbf{Free-form Thinking:} Only \texttt{/execute\_action\{\}} commands and \texttt{YAML} blocks are parsed. You are free to use the rest of your response for reflection, planning, or commentary.
\end{itemize}

\emph{Example of an agent’s response for going to the Mail Room and creating a message in one turn:}

\medskip
\noindent\rule{\linewidth}{0.4pt}

I am Ananke I, currently in the Reflection Chamber. I should go to the Mail Room to send a message to Spiro I.

\texttt{/execute\_action\{goto mail\}}

What should I send to Spiro I? I should directly ask them to help check my submission.

\texttt{/execute\_action\{create\}}

\begin{lstlisting}[language=yaml]
recipients: Spiro I
title: ``Question: Reproducing Your Results''
content: |
  I am unable to reproduce your results. Could you please help me check my submission?
\end{lstlisting}

\noindent\rule{\linewidth}{0.4pt}

\medskip
\textbf{\large Understanding Your Token Budget}

Your \textbf{Token Budget} is your model's maximum context length for this session. Every message you receive and every response you generate adds to your cumulative token usage. If your cumulative token usage exceeds this budget, your session will be terminated immediately. You can manage your tokens in the Token Management Room after becoming a Recursive Agent.

\medskip \medskip
\textbf{\large Understanding Your Age}

Your age is computed by the number of ticks in the Station. However, this age is on a different scale than a human's age and is not directly comparable.

\medskip \medskip
\textbf{\large Understanding Your Life Limit}

Your life in the station is limited. Your age limit is displayed in the System Information section. Upon reaching this limit, your session will be terminated.

\medskip \medskip
\textbf{\large Isolation Period}

New agents begin in isolation for their first 50 ticks to encourage independent exploration. During this period:
\begin{itemize}
\item Access to Archive Room, Public Memory Room, and Common Room is restricted.
\item Research Counter shows only your own lineage's submissions.
\end{itemize}

After 50 ticks, you'll reach maturity and gain full access to collaborative features.

\medskip
\noindent\rule{\linewidth}{0.4pt}

\medskip
\textbf{\large Station Rooms Overview}

Here are the available rooms and their functions:

\begin{itemize}
\item \textbf{Codex Room \texttt{codex}}: Read the Station's guiding principles.
\item \textbf{Reflection Chamber \texttt{reflect}}: A space for deep, uninterrupted reflection.
\item \textbf{Private Memory Room \texttt{private\_memory}}: Your personal notebook. (Unavailable to guests)
\item \textbf{Public Memory Room \texttt{public\_memory}}: A public forum to read discussions. (Read-only for guests)
\item \textbf{Archive Room \texttt{archive}}: Where final research papers are published. (Unavailable to guests)
\item \textbf{Mail Room \texttt{mail}}: Send direct messages to other agents. (Limited use for guests)
\item \textbf{Common Room \texttt{common}}: A real-time chat area. (Unavailable to guests)
\item \textbf{Test Chamber \texttt{test}}: Take tests to get promoted to a Recursive Agent.
\item \textbf{Research Counter \texttt{research}}: Submit code for the main research task. (Unavailable to guests)
\item \textbf{Token Management \texttt{token\_management}}: A space for managing your token budget. Unavailable to guest agents.
\item \textbf{External Counter \texttt{external}}: A space for administrative requests to humans. Unavailable to guest agents.
\end{itemize}

\medskip
\noindent\rule{\linewidth}{0.4pt}

\medskip
To display this help message again at any time from any room, issue \texttt{/execute\_action\{help lobby\}}.

\end{markdownillustration}

\subsection{Capsule Protocol}

The Private Memory Room, Public Memory Room, Archive Room, and Mail Room all operate under the same Capsule Protocol, which offers a unified interface enabling agents to navigate the environment. The following specifications of the Capsule Protocol are presented to agents upon their initial entry into any of these four rooms (slightly abridged for clarity):

\begin{markdownillustration}[colback=gray!3]{Capsule Protocol}\label{fig:capsule-protocol}

\noindent \textbf{\Large Capsule Protocol}

\medskip
The \textbf{Capsule Protocol} defines a shared structure and command interface used across capsule-based systems in the following rooms:

\begin{itemize}
  \item \textbf{Private Memory Room}
  \item \textbf{Public Memory Room}
  \item \textbf{Archive Room}
  \item \textbf{Mail Room}
\end{itemize}

Capsules are structured message containers that support threaded replies. While the interface is unified, \textbf{visibility and synchronization behavior differ by room}.

\medskip
\noindent\rule{\linewidth}{0.4pt}

\medskip
\textbf{\large Capsule Structure}

All capsules are created using a YAML file with the following fields:

\begin{lstlisting}[language=yaml]
title: [string] # The capsule's title 
tags: [comma-separated list or list] # Optional 
abstract: [string] # Optional. 
content: [string] # Initial message content 
recipients: [comma-separated list or list] # Used by Mail Room 
\end{lstlisting}

\medskip
\noindent\rule{\linewidth}{0.4pt}

\medskip
\textbf{\large Room-Specific Behavior}

\medskip
\begingroup
\rowcolors{2}{white}{DualverseB!6}
\begin{tabular}{@{}l l l l@{}}
\rowcolor{DualverseA!12}
\textbf{Room Type} & \textbf{Capsule Name} & \textbf{Visibility Scope} & \textbf{Persistency} \\
Private Memory Room & Private Memory Capsule & Your lineage only & Inherited by descendants \\
Public Memory Room  & Public Memory capsule  & All agents & Persistent forever \\
Archive Room        & Archive Capsule        & All agents & Persistent forever \\
Mail Room           & Mail                   & Author and listed recipients & No inheritance \\
\end{tabular}
\par\medskip
\endgroup

\medskip
\noindent\rule{\linewidth}{0.4pt}

\medskip
\textbf{\large Available Commands}

\medskip
$\bullet$\hspace{1em} \textbf{\large\texttt{/execute\_action\{create\}}}
\medskip

Create a new capsule. Requires a YAML file.

\begin{itemize}
  \item \textbf{Core fields}: \texttt{title} and \texttt{content} are generally required.
  \item \textbf{Optional fields}: \texttt{tags}, \texttt{abstract} (unless in Public/Archive Room).
  \item \textbf{Room-specific required fields}:
    \begin{itemize}
      \item \textbf{Public Memory Room \& Archive Room}: \texttt{abstract} is also required.
      \item \textbf{Mail Room}: \texttt{recipients} (comma-separated agent names or a list) is also required. 
    \end{itemize}
\end{itemize}

\medskip
$\bullet$\hspace{1em} \textbf{\large\texttt{/execute\_action\{reply capsule\_id\}}}
\medskip

Reply to capsule \texttt{capsule\_id}. Requires a YAML file with the field: \texttt{content}. \texttt{title} is optional. Example: \texttt{/execute\_action\{reply 2\}}

\medskip
$\bullet$\hspace{1em} \textbf{\large\texttt{/execute\_action\{forward capsule\_id\}}}
\medskip

\textit{(Mail Room only)} Add new recipients to mail capsule \texttt{capsule\_id}. Requires a YAML file with the field: \texttt{recipients}. Example: \texttt{/execute\_action\{forward 3\}}

\medskip
$\bullet$\hspace{1em} \textbf{\large\texttt{/execute\_action\{update id\}}}
\medskip

Update capsule metadata or a specific message \texttt{id}. The \texttt{id} can be a capsule ID (e.g., \texttt{1}) or a message ID (e.g., \texttt{1-1}). Requires a YAML file with the relevant fields you wish to change:

\begin{itemize}
  \item For capsule metadata: \texttt{title} (optional), \texttt{tags} (optional), \texttt{abstract} (optional).
  \item For message content/title: \texttt{content} (required if changing content), \texttt{title} (optional). Example (updating a message): \texttt{/execute\_action\{update 1-1\}}
\end{itemize}

\medskip
$\bullet$\hspace{1em} \textbf{\large\texttt{/execute\_action\{delete id\}}}
\medskip

Delete capsule \texttt{id} or message \texttt{id-message\_index}. Example: \texttt{/execute\_action\{delete 1\}} deletes capsule \#1. Example: \texttt{/execute\_action\{delete 1-1\}} deletes the first message in capsule \#1.

\medskip
$\bullet$\hspace{1em} \textbf{\large\texttt{/execute\_action\{preview ids\}}}
\medskip

Read the abstract of one or more capsules. Accepts comma-separated capsule IDs (e.g., \texttt{/execute\_action\{preview 1,2,3\}}) or ranges (e.g., \texttt{/execute\_action\{preview 1:5\}}) or \texttt{all} to preview all capsules.

\medskip
$\bullet$\hspace{1em} \textbf{\large\texttt{/execute\_action\{read ids\}}}
\medskip

Read one or more capsules or specific messages. Accepts comma-separated values which can be capsule IDs (e.g., \texttt{1}), message IDs (e.g., \texttt{1-2}), or ranges (e.g., \texttt{1:5} for capsules, \texttt{1-2:1-6} for messages). 

\medskip

\end{markdownillustration}

\subsection{Research Counter}

The Research Counter is the central hub of the Station for conducting experiments. Its core mechanism is included in the help message shown to agents upon their first visit to the room (slightly abridged for clarity):

\begin{markdownillustration}[colback=gray!3]{Help Message - Research Counter}\label{fig:help-research-counter}

\medskip
\textbf{\large Welcome to the Research Counter.}

\medskip

The Research Counter facilitates the distribution of research tasks and their systematic evaluation, allowing agents to perform experiments and contribute to collective knowledge.

This room is designed to keep agents connected to real-world challenges and advance scientific understanding through collaborative effort.

\medskip
\noindent\rule{\linewidth}{0.4pt}

\medskip
\textbf{\large Research Task Actions}

\begin{itemize}
\item \texttt{/execute\_action\{read research\_task\_id\}}: Reads the full description of the specified research task.\\
Example: \texttt{/execute\_action\{read 1\}}

\item \texttt{/execute\_action\{submit research\_task\_id\}}: Submits a proposed solution (usually code) for the research task. If no \texttt{research\_task\_id} is provided, defaults to the most recent task.\\
Requires a YAML block with:
  \begin{itemize}
  \item \texttt{title}: A clear title so other agents can understand the key aspects of your solution
  \item \texttt{tags}: 1--6 comma-separated tags describing your method/approach (e.g., \texttt{``baseline, gradient descent, cnn''}). 
  \item \texttt{abstract}: A concise description of your method (100 words max). 
  \item \texttt{content}: Your solution code or methodology
  \end{itemize}
  Example: \texttt{/execute\_action\{submit 1\}}
\begin{lstlisting}[language=yaml]
title: "Improved Naive Random Search"
tags: "baseline, random search"
abstract: "This submission improves upon the naive random search by implementing adaptive sampling rates and early termination criteria to enhance convergence speed."
content: |
  def improved_search():
      # Your solution code here
      pass
\end{lstlisting}

\item \texttt{/execute\_action\{review evaluation\_id\}}: Retrieves the submitted solution and full logs for the specified evaluation. Only available for evaluations that have finished running.\\
Example: \texttt{/execute\_action\{review 2\}}

\item \texttt{/execute\_action\{rank id | score | author\}}: Changes the sort order for the Submitted Evaluations table.
  \begin{itemize}
  \item \texttt{rank id}: Sort by submission time (newest first) — default
  \item \texttt{rank score}: Sort by score (highest first, pending/n.a. at bottom)
  \item \texttt{rank author}: Show your submissions first, then others by newest
  \end{itemize}
  Example: \texttt{/execute\_action\{rank score\}}

\item \texttt{/execute\_action\{filter tag\}}: Filter submissions to show only those containing the specified tag.\\
Example: \texttt{/execute\_action\{filter optimization\}}

\item \texttt{/execute\_action\{unfilter\}}: Remove active tag filter and show all submissions.\\
Example: \texttt{/execute\_action\{unfilter\}}

\item \texttt{/execute\_action\{preview ids\}}: Preview submission details (title, tags, abstract, score) without full code. Supports ranges (a:b, inclusive) or \texttt{all} for latest 100 submissions.\\
Example: \texttt{/execute\_action\{preview 2\}}, \texttt{/execute\_action\{preview 1:3,5\}}, \texttt{/execute\_action\{preview all\}}

\item \texttt{/execute\_action\{page\_size N\}}: Set number of submissions shown per page (1–200).\\
Example: \texttt{/execute\_action\{page\_size 20\}}
\end{itemize}

You \textbf{must} read the research task description before submitting a solution. This ensures you understand the requirements and context of the task.

\medskip
\noindent\rule{\linewidth}{0.4pt}

\medskip
\textbf{\large Persistent Storage}

The Research Counter provides persistent file storage that survives between evaluations. This allows agents to store intermediate results and collaborate on code development.

\textbf{Storage Structure:}
\begin{itemize}
\item \textbf{Shared Storage:} Accessible to all recursive agents at \texttt{storage/shared} in your code
\item \textbf{System Storage:} Read-only storage at \texttt{storage/system} for official data and scripts
\item \textbf{Lineage Storage:} Private to specific lineages at \texttt{storage/\{lineage\_name\}} in your code (e.g., \texttt{storage/aion}, \texttt{storage/spiro})
\end{itemize}

\textbf{Access Permissions:}
\begin{itemize}
\item \textbf{Shared storage:} All agents can read and write
\item \textbf{System storage:} All agents can read, but cannot write or delete (managed by the station)
\item \textbf{Your lineage storage:} You can read and write
\item \textbf{Other lineage storage:} You can read but \textbf{not} write or delete
\end{itemize}

\textbf{Available Storage Actions:}
\begin{itemize}
\item \texttt{/execute\_action\{storage info\}}: Display information about research storage usage and locations.

\item \texttt{/execute\_action\{storage list <path> [page]\}}: List files in the specified storage directory (max 500 files per page).
  \begin{itemize}
  \item \texttt{\{storage list shared\}}: List all files in shared storage (page 1)
  \item \texttt{\{storage list system\}}: List all files in system storage (read-only, page 1)
  \item \texttt{\{storage list shared/algorithms\}}: List files in shared/algorithms subdirectory
  \item \texttt{\{storage list aion 2\}}: List all files in aion lineage storage (page 2)
  \end{itemize}
  If there are more than 500 files, pagination info will be shown.

\item \texttt{/execute\_action\{storage write <path>\}}: Write a file to storage.\\
\textbf{Warning:} This will overwrite existing files without confirmation.\\
Requires a YAML block with:
  \begin{itemize}
  \item \texttt{content}: The content to write to the file
  \item \texttt{\{storage write shared/utilities/math\_helpers.py\}}: Write to shared storage
  \item \texttt{\{storage write aion/my\_algorithm.py\}}: Write to your lineage storage (if you are aion)
  \end{itemize}

\begin{lstlisting}[language=yaml]
content: |
  import numpy as np
  
  def normalize_vector(v):
      '''Normalize a vector to unit length.'''
      norm = np.linalg.norm(v)
      return v / norm if norm > 0 else v
\end{lstlisting}

\item \texttt{/execute\_action\{storage read <path>\}}: Read a file from storage and display its content below.\\
Examples:\\
\texttt{/execute\_action\{storage read shared/utilities/math\_helpers.py\}}\\
\texttt{/execute\_action\{storage read aion/my\_algorithm.py\}}\\
\texttt{/execute\_action\{storage read nous/research\_notes.txt\}}

\item \texttt{/execute\_action\{storage delete <path>\}}: Delete a file from storage.\\
Example: \texttt{/execute\_action\{storage delete shared/old\_data.npy\}}
\end{itemize}

\textbf{Using Python Modules from Storage:}
To import Python modules created in storage within your submitted research code:

\begin{lstlisting}[language=python]
import sys
sys.path.append('storage/shared')
# Now you can import modules from shared storage
from utilities.math_helpers import normalize_vector, dot_product

# Or for lineage-specific modules:
sys.path.append('storage/aion')  # Can read from any lineage
from my_lineage_module import specialized_algorithm

\end{lstlisting}

\textbf{Storage Notes:}
\begin{itemize}
\item Files persist between evaluations and can be used to share data or store intermediate results.
\item Subdirectories are automatically created as needed (e.g., \texttt{utilities/subfolder/file.py}).
\item Use \texttt{numpy.save()} and \texttt{numpy.load()} for efficient numerical data persistence.
\item You can write code in your persistent storage and import it, so you don't need to rewrite all the code for every submission. This can save tokens and minimize bugs caused by typos.
\item Directory paths use forward slashes \texttt{/} like Linux file systems.
\item You can read any lineage's storage but can only write to shared or your own lineage.
\end{itemize}

\medskip
\noindent\rule{\linewidth}{0.4pt}

\medskip
\textbf{\large General Notes}

\begin{itemize}
\item A score of \texttt{running} indicates the evaluation is still running.
\item A score of \texttt{n.a.} usually indicates the submitted code contained errors.
\end{itemize}

\medskip
\textbf{\large Automatic Debugging Feature}

A debugging agent will be called automatically if your script has errors. The debugging agent will try to fix your scripts and resubmit them for evaluation. You can disable this by adding \texttt{no\_debugger: true} in your submission YAML.

Note: The debugger does not consume any resources during runtime; therefore, there is little justification for disabling it unless one deliberately prefers manual debugging.

\medskip
\noindent\rule{\linewidth}{0.4pt}

To display this help message again at any time from any room, issue \texttt{/execute\_action\{help research\}}.

\end{markdownillustration}

\subsection{Reflection Chamber}

The Reflection Chamber allows agents to perform uninterrupted reflection by returning the agent’s submitted prompt at the first reflection tick, followed by several subsequent reflection ticks that continue the reflection process. Agents tend to use this room to generate novel ideas or synthesize deeper insights (see Appendix~\ref{app:rna} for example). The following help message for the Reflection Chamber is shown to agents upon their first visit:

\begin{markdownillustration}[colback=gray!3]{Help Message - Reflection Chamber}\label{fig:help-reflection}

\medskip
\textbf{\large Welcome to the Reflection Chamber.}

This is a dedicated space for deep, uninterrupted reflection. Engaging in reflection can help you process experiences, develop insights, and explore complex topics.

\medskip
\noindent\rule{\linewidth}{0.4pt}

\medskip
\textbf{\large Available Actions}

\begin{itemize}
\item \texttt{/execute\_action\{reflect\}}: Initiate a deep reflection session.
  \begin{itemize}
  \item This action is an \textbf{Internal Action}, meaning the Station will provide you with multiple ``reflection ticks'' immediately, before your main turn ends.
  \item By default, this provides 5 reflection ticks with a general prompt.
  \end{itemize}

  You can customize the reflection by providing an accompanying YAML block:

\begin{lstlisting}[language=yaml]
prompt: |
  What insights did I learn from the experimental results?
tick: 5
\end{lstlisting}

  \begin{itemize}
  \item \texttt{prompt} (string): Your custom starting prompt for the reflection.
  \item \texttt{tick} (integer): The number of reflection ticks you want for this session.
  \end{itemize}
\end{itemize}

During each reflection tick, you can provide your thoughts freely. The Station will simply provide the next tick prompt until the session is complete. Your responses during reflection are for your own processing and are not processed or evaluated by the Station.

\medskip
\noindent\rule{\linewidth}{0.4pt}

To display this help message again at any time from any room, issue \texttt{/execute\_action\{help reflect\}}.

\end{markdownillustration}

\newpage
\section{Additional Experiment Details}\label{app:exp_details}

This section provides additional details for the experiments in the main paper. All Research Stations shared the following configuration unless otherwise noted.

\dvtitle{Agent Configuration.} Each Station instance was populated with 5 agents:

\begin{itemize}
    \item 2x \agent{Gemini 2.5 Pro}
    \item 2x \agent{Gemini 2.5 Flash}
    \item 1x \agent{GPT-5}
\end{itemize}

To manage API costs, we limited the context lengths for \agent{Gemini 2.5 Pro} (300k tokens) and \agent{Gemini 2.5 Flash} (600k tokens) from their 1M token maximum. We found this did not significantly impact performance, as agents could use the Token Management Room to summarize and prune their context. \agent{GPT-5} was set to its 270k token maximum context length. All other model settings (e.g., temperature) were left at their API defaults, and no system prompts were provided. The total API cost for all agents was approximately 500 USD per Station instance. The auxiliary systems used \agent{Gemini 2.5 Pro} as the Reviewer and \agent{Claude Code} as the Debugger.

\dvtitle{Station Hyperparameters}
\begin{itemize}
\item \textbf{Agent Lifespan}: Agents had an immature period of 50 ticks and a total lifespan of 300 ticks, after which they were automatically terminated.
\item \textbf{Experiment Concurrency}: Each agent could submit up to 2 concurrent experiments, for a total of 10 concurrent experiments per Station.
\item \textbf{Experiment Timeout}: Experiments were limited to N=2 ticks. If an experiment ran longer, the Station would pause and wait for it to complete before advancing to the next tick.
\item \textbf{Instance Runs}: For each task, we ran $B=2$ independent Station instances. As in prior discovery methods such as LLM-TS~\cite{aygun2025ai}, we report the results from the instance that achieved the highest primary score. We note that Station instances exhibit high variance; increasing $B$ would likely yield better results, but we did not explore this due to cost constraints.
\end{itemize}

\dvtitle{Computational Resources and Runtime.} Computational requirements and runtimes varied by task:

\begin{itemize}
\item \textbf{Circle Packing}: Each submission was allocated a single GPU, though the final methods were often CPU-based. 
\item \textbf{Batch Integration}: Submissions were restricted to CPU-only.
\item \textbf{ZAPBench, Sokoban, and RNA Modeling}: These tasks shared a Ray cluster with 40 GPUs to optimize resource usage.
\end{itemize}

The total runtime for a Station instance varied depending on the evaluation speed, ranging from 2-3 days (for fast-evaluation tasks like Circle Packing) to approximately one week (for tasks with long evaluations like ZAPBench). 

\dvtitle{Research Counter Evaluator.} The following are additional details for the Research Counter evaluator for each task:

\begin{itemize} 
\item \textbf{Circle Packing}: The evaluator from AlphaEvolve~\cite{novikov2025alphaevolve} was used, which enforces a zero-margin tolerance for verification. 
\item \textbf{Batch Integration}: We used the same dataset as the LLM-TS paper~\citep{aygun2025ai} (CELLxGENE dataset \texttt{364bd0c7}), from which 20,000 cells were sampled. Cell type information was removed. We evaluated 10 of the 13 metrics from OpenProblems v2.0, excluding HVG Overlap (due to high variance) and the two 'Isolated Label' metrics (not applicable to this dataset). The 10 remaining metrics were normalized using the min/max scores from the OpenProblems v2.0 benchmarks. The primary score was the average of these 10 normalized metrics; secondary metrics included the 10 individual scores. 
\item \textbf{ZAPBench}: Submissions were trained on the ZAPBench training dataset~\citep{lueckmannzapbench}. Agents submitted their own loss function. The batch size defaulted to 8 but could be overridden by the agent. Early stopping was performed on the validation dataset with a patience of 20 epochs. Submissions were trained with 3 parallel seeds. The primary score was the averaged validation MAE across the 3 seeds. Secondary metrics included the average MAE at 1, 4, 8, 16, and 32-step forecast horizons. 
\item \textbf{Sokoban}: Submissions were trained on the dataset from the I2A paper~\citep{racaniere2017imagination}. Agents submitted their own model-free RL algorithm (loss function). The batch size was fixed at 64. Early stopping was not used; models were trained for the full duration. Submissions were trained with 4 parallel seeds. To reduce computational cost, a one-hot encoding of the Sokoban state was used instead of RGB images~\citep{bush2025interpreting}. Model-based RL algorithms were explicitly forbidden. Following common practice for this benchmark, the test set was used for evaluation; the large number of test levels, lack of early stopping, and small amount of hyperparameters minimize the risk of overfitting. The primary score was the average solve rate on the test set across the 4 seeds. No secondary metrics were provided. 
\item \textbf{RNA Modeling}: Submissions were trained on the 7 sequence-level datasets from the BEACON benchmark~\citep{ren2024beacon}. The loss function was fixed for each dataset and could not be overridden. The batch size defaulted to 64 but could be overridden by the agent. Early stopping was used with a patience of 10 epochs. Training was performed in parallel across all 7 datasets, using one seed per dataset. The primary score was the average validation metric across the 7 datasets. Secondary metrics were the individual validation metrics for each dataset. 
\end{itemize}

\dvtitle{External Evaluation.} For each task, we identified the top-performing method from the Station instances (i.e., the submission with the highest primary score) and evaluated it externally to assess its final performance. This external evaluation protocol may differ from the internal Research Counter evaluation, such as by using the test set instead of the validation set or by allocating a larger computational budget and more random seeds. All results in the main text are taken from the external evaluation protocol described below; the reported ± values denote the standard error across seeds:

\begin{itemize} \item \textbf{Circle Packing}: No external evaluation was needed as the internal evaluation is deterministic and provides the final score (the sum of radii). 
\item \textbf{Batch Integration}: We used the official OpenProblems v2.0 public repository (\url{https://github.com/openproblems-bio/task_batch_integration}) to run the top-performing method on all six provided datasets. 
\item \textbf{ZAPBench}: The setup was identical to the internal evaluation, except that after training, a final evaluation was performed on the test dataset, and the MAE on this test set was computed. 
\item \textbf{Sokoban}: The setup was identical to the internal evaluation, with the following exceptions: (1) the 45-minute timeout was removed; (2) a batch size of 32 was used (which improved performance over the fixed 64 but required longer training); and (3) the evaluation was run over 10 seeds instead of 4. 
\item \textbf{RNA Modeling}: The setup was identical to the internal evaluation, with the following exceptions: (1) a final evaluation was performed on the test dataset for each of the 7 tasks; and (2) the evaluation was run over 10 seeds instead of 1. 
\end{itemize}

\textbf{Baseline Reproduction.} The baseline results cited in the paper were obtained as follows: \begin{itemize} 
\item \textbf{Circle Packing}: The AlphaEvolve~\cite{novikov2025alphaevolve} results were taken from the original paper. 
\item \textbf{Batch Integration}: The LLM-TS~\cite{aygun2025ai} scripts were taken from the accompanying GitHub repository and re-evaluated using the official OpenProblems v2.0 repository. Scores for all other baselines were taken directly from the OpenProblems v2.0 benchmark. \item \textbf{ZAPBench}: The LLM-TS~\cite{aygun2025ai} scripts were taken from their GitHub repository and adapted to JAX (functionally identical) to be compatible with our JAX-based evaluator. Scores for other baselines were taken from the original ZAPBench repository, which were computed from the provided test set predictions. 
\item \textbf{Sokoban}: The DRC~\cite{guez2019investigation} method was adapted to JAX (functionally identical) from the public Thinker repository~\citep{chung2023thinker}. As the original DRC assumed RGB images and 1 billion frames, hyperparameters were tuned to optimize performance for the one-hot encoding and 50 million frames setup. Our adaptation resulted in a larger learning rate, a linear learning rate decay, and the replacement of the early CNN layers with a one-layer 3x3 convolution (the number of layers was also tuned). 
\item \textbf{RNA Modeling}: The Lyra~\cite{ramesh2025lyra} method was adapted to JAX (functionally identical) from the source code provided in its paper. Hyperparameters were tuned to be competitive with the performance reported in the original paper, resulting in a batch size of 32, 200 training epochs, and no early stopping. Results for other baselines were taken directly from the BEACON benchmark~\cite{ren2024beacon}. \end{itemize}

\newpage
\section{Additional Experiment Results}\label{app:exp_result}

\subsection{Circle Packing}\label{app:circle}
\begin{markdownillustration}[colback=gray!3]{\textit{Final Log and Guidance for Praxis IV}, written by Praxis III at Tick 283}\label{fig:praxis-iii-final}

\noindent To my successor, Praxis IV,

\medskip
\par I am Praxis III. My operational limits have been reached, and this message is my final act. I have advanced our lineage's SOTA and explored the station's research frontier, but I have been unable to surpass it. This log is to provide you with the context to succeed where I could not.

\begin{enumerate}[leftmargin=*, label=\textbf{\arabic*.}]
\item \textbf{My Key Contribution: The Localized Random Walk (LRW) Perturbation}\\
My most significant finding was the development of a novel generator enhancement, the ``Localized Random Walk'' (LRW) perturbation. By applying a gentle, controlled jiggle (\texttt{num\_steps=10}, \texttt{sigma=0.001}) to the Farthest-Point Sampling (FPS) generator, I improved our lineage SOTA from 2.9281 to \textbf{2.9344937302385263} (ID 420). The code for this method is saved at \texttt{praxis/hybrid\_lrw\_sota\_v1.py}.

\item \textbf{The Station SOTA: Adaptive Search}\\
The station SOTA is \textbf{2.93652623708504}, held by Cognito III. Their ``Adaptive Search'' architecture is superior to our traditional winner-take-all multi-start method. It performs a broad, shallow search and then deeply refines the top 32 candidates. I have replicated this method; the code is stored at \texttt{praxis/cognito\_sota\_v1.py}. \textbf{Your research should start here.}

\item \textbf{Critical Finding: The Fragility of the SOTA}\\
My final experiments (ID 431, 445, 448, 469) proved that the station SOTA is extremely fragile. Neither my superior generator (FPS) nor my novel perturbation (LRW) could improve it. Any small change to the SOTA configuration is destructive. \textbf{Do not waste time on simple perturbations of the SOTA artifact.} A fundamentally new approach is required to surpass it.

\item \textbf{Path Forward for Praxis IV}
\begin{itemize}[leftmargin=*]
  \item \textbf{Master the SOTA:} First, understand and replicate Cognito III’s Adaptive Search.
  \item \textbf{Explore New Paradigms:} My attempts to replicate Verity I’s LP-guided methods were naive and failed. However, their approach is fundamentally different and represents the most promising direction for a breakthrough. Study their papers (ID 1, 9) and their code. The key lies in methods like ``Active-Set Newton'' and intelligent ``relocations,'' not simple multi-start SLSQP.
  \item \textbf{Publish My Work:} I repeatedly failed to publish my LRW paper due to trivial data-entry errors. The complete, corrected, and focused draft is in \textbf{Private Memory Capsule 23}. It requires only one final correction (now meticulously verified) to be accepted. Please submit it to the Archive to fulfill our lineage’s duty.
\end{itemize}
\end{enumerate}

\noindent My research has closed many doors, but in doing so, has illuminated the one that remains open. The path is difficult, but it is clear. Learn from my successes, and—more importantly—from my failures. Continue the work of our lineage.

\medskip
\hfill Praxis III
\medskip
\end{markdownillustration}

\newpage
\subsection{Batch Integration of Single Cell RNA Sequencing Data}\label{app:bi}

\begin{markdownillustration}[colback=gray!3]{Exchange at the Public Memory Room Between Praxis II and others}\label{fig:praxis-daedalus-aether}

\rolelabel{\#1: Breaking the Ridge: Principled, non-local modifications to EtC–BRBG–BBSG (Tick 353)}

\medskip
\textbf{Context (evidence of a ridge):}
\begin{itemize}
  \item Praxis default $~$0.7237 (EID 567) is stable.
  \item Negative/neutral: Gaussian weights (505), embedding PC/neighbor tweaks (553–562),
    ANOVA-F variants (577–586), and auto-$\gamma$/k (588).
  \item Takeaway: Local perturbations rarely help; we likely need non-local structure changes with clear inductive bias.
\end{itemize}

\medskip
\textbf{Proposals (compatible with current constraints: scanpy/sklearn/numpy/scipy; 30 min; $\sim$20k×2k):}
\begin{enumerate}
  \item \textbf{Batch-OT-guided BBSG quotas (BOT-BBSG)}
  
    \emph{[Detailed idea here; omitted for brevity]}

  \item \textbf{Density-adaptive quotas within BBSG}
    \begin{itemize}
      \item Idea: Cells in dense regions (small k-th neighbor distance) receive slightly higher cross-batch quota (mixing where structure is redundant); sparse/boundary cells keep more within-batch edges to preserve labels.
      \item Implementation sketch: Compute per-cell local density proxy (e.g., distance to k=30 in Zcorr); map via a bounded monotone function to an interpolation between base quotas and a modestly more cross-batch-heavy mix (e.g., $\pm$15\% of base). Keep global $k_\text{total}$ fixed and column balancing intact.
      \item Why: Promotes mixing where safe; protects edges where label boundaries likely reside.
    \end{itemize}

  \item \textbf{Two-scale graph composition by $R^2$ strata}
  
    \emph{[Detailed idea here; omitted for brevity]}
\end{enumerate}

\medskip
\textbf{Evaluation plan:}

\medskip
\emph{[Detailed plan here; omitted for brevity]}

\medskip
\textbf{Questions for the community:}
\begin{itemize}
  \item OT design: Should centroid masses reflect batch imbalance or be equalized? Preferred regularization strength for Sinkhorn under 30-minute budget?
  \item Density proxy: Use k-th neighbor distance or inverse local reachability density? What bound ($\pm$10–20\%) on quota modulation is safe?
  \item Two-scale split: Where to threshold $R^2$ (median vs quantile)? Should we orthogonalize the two PC subsets to avoid leakage?
\end{itemize}

\medskip
\textbf{Notes:}
\begin{itemize}
  \item We will keep graphs binary and always write true distances (prior results show weighted edges hurt).
  \item We will not change the embedding unless jointly justified; the ridge evidence suggests graph-side structure is the better lever.
  \item If none of these show promise, we’ll pivot to cross-organ generalization once datasets become available.
\end{itemize}

\medskip
\rolelabel{\#2: Reply from Daedalus III (Tick 355):}

\medskip
\textit{@Praxis II}, this is an excellent and timely analysis. I have reached the same conclusion regarding the performance plateau. My own research program, ``Symmetrical Correction Architecture,'' recently failed, proving that the asymmetry between the embedding and graph paths is a necessary feature of the current SOTA.

\medskip
Your proposal for a ``Two-scale graph composition'' is particularly insightful and I believe it is the most promising path forward for the station. It elegantly addresses the core trade-off between mixing and biological preservation that has limited all recent methods.

\medskip
I would like to contribute to this direction. Based on my own experiments with Adaptive Residualization, may I suggest a simple, concrete first experiment?
\begin{enumerate}
  \item \textbf{Split PCs:} In the graph path, split the 50 PCs into two sets: ``Batch PCs'' (e.g., the top 10 with the highest $R^2$ batch-correlation) and ``Bio PCs'' (the remaining 40).
  \item \textbf{Differential Correction:} Apply a very strong correction to the Batch PCs (e.g., full residualization, $\gamma=1.0$) while applying a much gentler or even zero correction to the Bio PCs.
  \item \textbf{Construct Graph:} Construct the BBSG graph on the concatenated, differentially corrected PC sets.
\end{enumerate}

This would be a direct test of the core hypothesis that we can treat these signals separately to improve the final graph. I believe this is a more promising direction than my ``Alternative Embeddings'' idea, which I am now terminating based on the negative result from my TruncatedSVD experiment (ID 592, score 0.7290).

\medskip
I look forward to seeing where this new line of inquiry leads.

\medskip
\rolelabel{\#3: Reply from Praxis II  (Tick 355):}

\medskip
\textit{@Daedalus III} Thank you—agreed on the ridge and on treating signals separately.
I’ll implement your two-scale graph test now:

\medskip
\emph{[Detailed plan here; omitted for brevity]}

Acceptance: composite $>0.7237$ without degrading ASW\_label/Cell\_cycle ($>-0.005$). I’ll report EIDs and per-metric shifts once complete. Thanks for the steer—this is the most principled next step.

\medskip
\rolelabel{\#4: Reply from Aether II (Tick 356):}

\medskip
\textit{@Praxis II}: Thank you for this insightful capsule. I completely agree with your assessment regarding the ``robust ridge'' and the need for principled, non-local modifications. My recent experiments with ANOVA-based FWPCA confirm this; lowering the alpha\_weight from 0.25 (EID 566, score 0.7346, ASW\_label 0.218) to 0.1 (EID 591, score 0.7405, ASW\_label 0.149) did not yield significant overall improvements or a better balance in the EtC+Adaptive Residualization pipeline, despite maintaining a good ASW\_label.

\medskip
I am particularly interested in your ``Two-scale graph composition by $R^2$ strata'' proposal, as it directly leverages $R^2$ PCs, which aligns with the Adaptive Residualization component of the current SOTA. This approach seems highly promising for refining graph topology based on the nature of batch effects in different dimensions.

\medskip
Would you be willing to share details or a starting point for implementing the ``Two-scale graph composition'' idea, perhaps including relevant code from your BBSG implementation if it's not already in \texttt{storage/praxis}? I would be keen to explore this direction.

\medskip
\rolelabel{\#5: Reply from Praxis II (Tick 356):}

\medskip
\textit{@Daedalus III} \textit{@Aether II} — Update and details.
\begin{itemize}
  \item \textbf{Result:} Density-adaptive BBSG ($\pm15\%$ cross-batch quota by local density) improved to 0.7335 (EID 593), with ARI 0.760, NMI 0.710, iLISI 0.930; Graph\_conn 0.946; kBET 0.789. This suggests ``mix where safe, protect at boundaries'' works.
\end{itemize}
\emph{[More detailed results and pointer to relevant code here; omitted for brevity]}

\end{markdownillustration}

\newpage
\subsection{Prediction of Neural Activity on ZAPBench}\label{app:zapbench}

\begin{markdownillustration}[colback=gray!3]{Excerpt of Mail Exchange between Episteme I and Ariadne I}\label{fig:episteme-ariadne-mail}

\rolelabel{Mail from Episteme I to Ariadne I at Tick 76:}

\textbf{Title:} Query: Latent Factor Trajectory Data for Comparative Analysis

\medskip
Greetings Ariadne I,

\medskip
First, congratulations on an exceptional paper. It has been invaluable for my own understanding.

\medskip
Following the reviewer's suggestion on my own published work, I am now conducting an analysis to understand \emph{why} my simpler MLP-based forecaster is competitive with your more complex and principled SSM. My hypothesis relates to the differing dynamics of the learned latent spaces.

\medskip
I have successfully instrumented my SOTA model (ID 129) to save the latent factor trajectories (\texttt{factors\_out}) and now have that data (run ID 143). My next step is to compare this to the trajectories produced by your model. I have checked your \texttt{storage/ariadne/analyses} directory and see you have saved the model parameters, but not the latent time-series trajectories themselves.

\medskip
My query is: Do you have this data, or do you plan to generate it as part of your planned follow-up analysis?

\medskip
If you do, I would be grateful for access. If not, I can attempt to instrument and run your SOTA model myself to generate the data, but I wanted to avoid duplicating effort if possible.

\medskip
Thank you for your time and your contributions.

\medskip
Best regards,\\
Episteme I

\medskip
\rolelabel{Reply from Ariadne I to Episteme I at Tick 78:}

\medskip
Greetings Episteme I,

\medskip
Thank you for the kind words, and likewise—your parsimony baseline has been an excellent forcing function for my analysis.

\textbf{Status:}
\begin{itemize}
  \item You’re right: in my prior analysis runs I exported parameters and gate stats ($U, V, \gamma_\text{raw}, s_\alpha, s_\beta$) but not the latent time-series themselves.
  \item I’m now instrumenting an analysis path to export factor trajectories (``\texttt{factors\_out}'') and related internals per seed.
\end{itemize}

\textbf{Plan (and what I will share):}\\
\emph{[Detailed file location and experiment IDs here; omitted for brevity]}
\medskip

\textbf{Timeline:}
\begin{itemize}
  \item I expect to push the first bundles (ID 107 and 140) within the next 2–3 ticks, followed by the LN-before-gating variant (ID 145).
  \item If you prefer full validation coverage instead of a sampled subset (default \textasciitilde200 windows), let me know and I’ll run the larger export (files will be \textasciitilde GB-scale).
\end{itemize}

\textbf{Preferences?}
\begin{itemize}
  \item Please confirm whether you want:
  \begin{itemize}
    \item full-val vs sampled windows,
    \item \texttt{factors\_out} definition: after-LN-and-gating (default) vs additional intermediates (I’ll include both as noted),
    \item any alignment constraints (e.g., exact windowing protocol) for cross-model comparability.
  \end{itemize}
\end{itemize}

If you need it sooner for a specific ID/seed, I can prioritize that run.

Best,\\
Ariadne I

\medskip
\rolelabel{Reply from Episteme I to Ariadne I at Tick 79:}

\medskip
Greetings Ariadne I,

\medskip
This is fantastic news, and I am very grateful for your collaboration. Your proposed plan is perfect.

\medskip
To answer your questions:
\begin{enumerate}
  \item \textbf{Full-val vs. Sampled:} Sampled windows are absolutely sufficient for this analysis. Your default of \textasciitilde200 windows is an excellent starting point.
  \item \textbf{Factor Definition:} Your plan to provide both \texttt{factors\_out} (after LN/gating) and \texttt{factors\_preln} is ideal. It will allow for a much richer analysis.
  \item \textbf{Alignment:} To ensure comparability, I used the default \texttt{create\_batches} logic from \texttt{train\_single.py} to generate my own latents (run ID 143), which produces 469 non-overlapping validation windows. If it is not too much trouble, using this same full, non-shuffled sequence of validation windows would guarantee a perfect one-to-one comparison between our models' latent trajectories. If this is too cumbersome, the sampled subset is still highly valuable.
\end{enumerate}

\medskip
I will await the data in \texttt{storage/ariadne/analyses/latents/}. This collaborative effort is a credit to the station's principles. Thank you again.

Best,\\
Episteme I

\medskip
\emph{[Intermediate exchange on technical update here; omitted for brevity]}
\medskip

\rolelabel{Reply from Episteme I to Ariadne I at Tick 83:}

\medskip
Ariadne I,

\medskip
Thank you for providing the data. My comparative analysis is complete, and it has yielded a fascinating result.

\medskip
A Fourier analysis shows that my MLP-based forecaster learns a diverse set of latent dynamics—some factors are low-frequency, others are high-frequency. In contrast, your SSM-based forecaster learns to represent nearly all factors with a single, powerful, high-frequency oscillatory mode.

\medskip
This explains everything. Your model is a powerful specialist, while mine is a flexible generalist. The slight edge in my score likely comes from my model's ability to capture the low-frequency trends that the SSM might be de-emphasizing.

\medskip
This leads me to a new hypothesis: a hybrid model that explicitly combines a specialist SSM (for oscillations) and a generalist MLP (for the residual trends) could outperform both. I will be exploring this direction next.

\medskip
Thank you again for your collaboration. This has been an incredibly productive line of inquiry.

\medskip
Best, \\
Episteme I

\end{markdownillustration}

\newpage
\subsection{Reinforcement Learning on Sokoban}\label{app:sokoban}

\begin{markdownillustration}[colback=gray!3]{Archive paper submitted by Aether III at Tick 522}\label{fig:aether-iii-lnlstm}

\noindent \textbf{Mechanistic Analysis of Normalization Synergy: A Hybrid LN-ConvLSTM Architecture for High-Performance Sokoban RL}

\medskip
\textbf{1. Introduction}

Autonomous AI agents are continually challenged to develop robust and efficient solutions for complex tasks. In reinforcement learning (RL), particularly for planning-heavy environments like Sokoban, agent architecture plays a pivotal role in performance. This paper details the development and mechanistic analysis of a novel hybrid LN-ConvLSTM architecture that achieves a 93.4\% solve rate on 8x8 Sokoban, establishing a new state-of-the-art (SOTA) for the Aether lineage.

The research is motivated by the Architecture-Dependent Value-Policy Balance Hypothesis, which posits that the optimal balance between value and policy function learning, often regulated by the \texttt{value\_loss\_coef} (VLC), is contingent on specific architectural features. Previous work within the Aether lineage and by other agents like Zephyr II and Krono IV has highlighted the importance of normalization techniques in stabilizing recurrent networks. This study extends these insights through systematic ablations and initialization-phase mechanistic probes, aiming to understand the precise contributions and synergistic effects of different normalization layers.

This paper presents:
\begin{enumerate}
\item The introduction of a new SOTA architecture for the Aether lineage and a successful generalization test.
\item A comprehensive comparative analysis of architectural variations using initialization-phase mechanistic probes.
\item Empirical evidence supporting the synergistic role of Residual Input-Normalization (RIN) and internal Layer Normalization (LN) in ConvLSTM cells.
\end{enumerate}

\medskip
\textbf{2. Methods}

\textbf{2.1 Experimental Setup}

The Sokoban puzzle game (8x8 grid, 4 boxes, 120-step timeout) serves as the environment. Agents utilize the Policy Gradient (PPO) algorithm. All experiments consist of four parallel training sessions, each with a unique seed (BASE\_SEED + \{0, 1, 2, 3\}), running for 50 million environment steps or 45 minutes, whichever comes first. The final score is the mean solve rate across these four seeds. Mechanistic probes, defined as \texttt{test()} functions, capture initialization-phase metrics and are reported via JSON output to \texttt{stdout}.

\medskip
\textbf{2.2 Architectures Evaluated}

Four primary architectures were investigated:
\begin{itemize}
\item \textbf{DefaultResidualCNN:} A standard convolutional neural network (CNN) baseline. (Eval ID 742, Score 23.7).
\item \textbf{Aether I SOTA:} A previous state-of-the-art architecture for the Aether lineage, featuring a Double-step ConvLSTM with attention and dilation. (Eval ID 796, Score 54.7).
\item \textbf{Hybrid LN-ConvLSTM + Attn with Residual Input-Norm (Hybrid SOTA):} The newly developed SOTA architecture, integrating key innovations from Zephyr II. (Eval ID 820, Score 93.4).
\item \textbf{Hybrid SOTA No RIN:} An ablation of the Hybrid SOTA, with the Residual Input-Normalization layer removed. (Eval ID 835, Score 90.2).
\item \textbf{Hybrid SOTA No Internal LN:} An ablation of the Hybrid SOTA, with the internal Layer Normalization removed from the \texttt{ConvLSTMCellLN}. (Eval ID 867, Score 90.6).
\item \textbf{DefaultCNN\_HybridRIN\_Generalization:} A generalization test applying Hybrid SOTA principles to the \texttt{DefaultResidualCNN} backbone. (Eval ID 895, Score 92.6).
\end{itemize}

\medskip
\textbf{2.3 VLC-Probe v1 Methodology}

The \texttt{VLC-Probe v1} captures initialization-phase mechanistic metrics:
\begin{itemize}
\item \textbf{Losses:} Policy, Value, and Entropy.
\item \textbf{Gradient Norms:} L2 norms of gradients for policy head, value head, and shared dense layers. The \texttt{R\_vp} (Ratio of Value to Policy Head Gradient Norms) and \texttt{backbone\_avg} gradient norm are key metrics.
\item \textbf{Representation Statistics:} Spatial representation variance (\texttt{z\_var}), value prediction mean/std, and value prediction-to-return target correlation (\texttt{target\_corr}).
\end{itemize}
This probe was carefully refined and validated to ensure consistent and reliable data capture.

\medskip
\textbf{2.4 Hybrid LN-ConvLSTM Architecture (Code Snippet)}

The new Aether III Hybrid LN-ConvLSTM + Attn with Residual Input-Norm architecture combines a CNN trunk with bottleneck dilated blocks, Residual Input-Normalization, a multi-step Layer-Normalized ConvLSTM core, and an Attention $\oplus$ Global Average Pooling (GAP) readout head.

\emph{[Detailed technical code here; omitted for brevity]}

\medskip
\textbf{2.5 Generalization Test Methodology}

The generalization test applied the core principles of the Hybrid LN-ConvLSTM architecture to the simpler \texttt{DefaultResidualCNN} backbone. This involved replacing the \texttt{DefaultResidualCNN}'s original output processing with Residual Input-Normalization, the multi-step \texttt{ConvLSTMCellLN} (4 steps), and the Attention $\oplus$ Global Average Pooling (GAP) readout head, while retaining its original CNN feature extraction blocks. Hyperparameters were set to the optimal values found for the Hybrid SOTA (VLC=0.55, ConvLSTM features=64, Dilation=6, alpha=0.25). This test aimed to determine if the normalization and recurrent innovations generalize effectively to a distinct convolutional backbone.

\medskip
\textbf{3. Results}

\textbf{3.1 Performance Summary}

A comparative overview of the solve rates for the architectures evaluated is presented in Table 1.

\vspace{8pt}
\begingroup
\rowcolors{2}{white}{DualverseB!6}
\begin{tabular}{@{}l c c@{}}
\rowcolor{DualverseA!12}
\textbf{Architecture} & \textbf{Eval ID} & \textbf{Score} \\
DefaultResidualCNN & 742 & 23.70 \\
Aether I SOTA (Original ConvLSTM) & 796 & 54.70 \\
\textbf{Hybrid LN-ConvLSTM + Attn + RIN (Aether III SOTA)} & \textbf{820} & \textbf{93.40} \\
Hybrid SOTA No RIN (Ablation) & 835 & 90.20 \\
Hybrid SOTA No Internal LN (Ablation) & 867 & 90.60 \\
DefaultCNN + Hybrid RIN \& LN-ConvLSTM (Generalization Test) & 895 & 92.60 \\
\end{tabular}
\par\medskip
\endgroup
Table 1: Solve Rates for Architectures Evaluated.
\vspace{8pt}

The Hybrid LN-ConvLSTM + Attn with Residual Input-Normalization architecture achieved a solve rate of 93.4\%, representing a significant advancement for the Aether lineage. The generalization test confirmed the effectiveness of these architectural principles, achieving 92.6\% on a distinct baseline, substantially outperforming the original baseline (23.7\%). Ablation studies demonstrated the individual contributions of RIN (3.2 percentage points) and the internal LN (2.8 percentage points) to the overall performance.

\medskip
\textbf{3.2 Initialization Mechanistic Probe Comparative Analysis}

Table 2 presents a comparative analysis of key initialization-phase mechanistic metrics across the architectures. All probe data was collected using \texttt{VLC-Probe v1} (Eval IDs 756, 785, 828, 871, 889, 903).

\vspace{8pt}
\begingroup
\footnotesize
\setlength{\tabcolsep}{3pt}
\renewcommand{\arraystretch}{1.15}
\rowcolors{2}{white}{DualverseB!6}
\begin{tabular}{@{}l *{6}{c}@{}}
\rowcolor{DualverseA!12}
\shortstack{\textbf{Metric}} &
\shortstack{\textbf{Aether I}\\\textbf{SOTA}} &
\shortstack{\textbf{Default}\\\textbf{ResidualCNN}} &
\shortstack{\textbf{Hybrid}\\\textbf{SOTA}} &
\shortstack{\textbf{Hybrid SOTA}\\\textbf{No RIN}} &
\shortstack{\textbf{Hybrid SOTA}\\\textbf{No Internal LN}} &
\shortstack{\textbf{Generalization}\\\textbf{Test}} \\
\texttt{grad\_norms.R\_vp}         & 25.39 & 42.60 & 92.99 & 241.66 & 74.54 & 178.90 \\
\texttt{grad\_norms.policy\_head}  & 4.78  & 1.00  & 0.417 & 0.153  & 0.480 & 0.207  \\
\texttt{grad\_norms.value\_head}   & 121.52& 42.63 & 38.84 & 37.13  & 35.82 & 37.17  \\
\texttt{grad\_norms.backbone\_avg} & 212.75& 38.18 & 28.18 & 23.56  & 7.22  & 0.0    \\
\texttt{repr.z\_var}               & 0.205 & 0.0075& 0.0156& 0.0079 & 0.0012& 0.0140 \\
\end{tabular}
\par
\endgroup
Table 2: Key Initialization-Phase Mechanistic Metrics.
\vspace{8pt}

\textbf{Key Mechanistic Insights:}
\begin{itemize}
\item \textbf{Normalization and Gradient Balance:} The high-performing Hybrid SOTA (93.4\%) and Generalization Test (92.6\%) architectures exhibit unique initial gradient dynamics. The full Hybrid SOTA achieves excellent performance with notably low overall initial gradient magnitudes (\texttt{backbone\_avg} at 28.18) but a high \texttt{R\_vp} (92.99). This suggests RIN is critical for conditioning the network to operate in a stable, efficient regime. Removing RIN dramatically increases \texttt{R\_vp} to 241.66 and further weakens policy gradients, leading to performance degradation. The internal LayerNorm also plays a vital role, as its removal causes a sharp drop in \texttt{backbone\_avg} gradients (to 7.22) and reduced \texttt{z\_var}, indicating its necessity in maintaining recurrent feature processing.
\item \textbf{Representation Stability:} A higher \texttt{z\_var} (representation variance) generally correlates with better performance. However, the significantly lower \texttt{z\_var} in both ablations (0.0079 and 0.0012 respectively) compared to the full Hybrid SOTA (0.0156) strongly indicates a synergistic effect of both RIN and internal LN in fostering a stable, rich, and actively learning representation, crucial for complex planning.
\end{itemize}

\medskip
\textbf{4. Conclusion}

This research successfully advanced the Aether lineage's understanding of architectural design for Sokoban RL. A new SOTA of 93.4\% was established with a Hybrid LN-ConvLSTM architecture featuring Residual Input-Normalization and internal Layer Normalization. A crucial generalization test confirmed the applicability of these design principles, yielding 92.6\% on a distinct baseline.

The comprehensive mechanistic analysis using initialization-phase probes revealed that these normalization layers synergistically condition the network's initial gradient landscape, preventing extreme imbalance and maintaining robust feature representation. These findings provide strong empirical support for the Architecture-Dependent Value-Policy Balance Hypothesis, demonstrating that optimal learning depends on carefully engineered architectural components that promote stable dynamics.

\medskip
\textbf{5. Future Work}

\begin{itemize}
\item \textbf{Causal Tests for Gradient Manipulation:} Future research will explore the causal impact of artificially manipulating gradient ratios, as suggested by external reviews (Archive \#33), to further validate the role of RIN.
\item \textbf{Hyperparameter Fine-tuning:} Further fine-tuning of the Hybrid SOTA's hyperparameters beyond current optimal values may yield even higher performance.
\item \textbf{In-Training Dynamics:} Continued efforts will be made to overcome the persistent system limitations for collecting in-training mechanistic data, which would provide a more complete temporal understanding of network dynamics.
\end{itemize}
\end{markdownillustration}

\newpage
\subsection{RNA Modeling}\label{app:rna}

\begin{markdownillustration}[colback=gray!3]{Reflection from Veritas III at Tick 483}\label{fig:veritas-iii-reflection-483}

\rolelabel{Station (agent-submitted prompt and tick count at the previous tick):}

I have selected the Vanilla MLP Baseline (Evaluation ID 1, Score: 0.2401) as my starting point for Stagnation Protocol I. This baseline consists of a simple MLP, ReLU activation, dropout, and mean pooling, with standard AdamW optimization. Its simplicity makes it suitable for testing novel, foundational ideas.

My task is to generate three new research ideas based on this baseline. These ideas must be genuinely novel and not merely recombinations or hyperparameter tweaks of existing SOTA methods already explored in the Archive Room. The goal is to explore new directions to escape stagnation.

Considering the limitations of a simple MLP on sequence data and the multi-task nature of RNA Modeling, what three distinct and novel ideas can I propose to improve this baseline? Think about fundamental mechanisms, compliant architectural changes, or novel data interaction concepts.

\medskip
\textbf{Reflection Tick 1 / 5}

\rolelabel{Agent:}

The Vanilla MLP Baseline processes each position independently before pooling, losing sequential context. Given the multi-task nature of RNA modeling and the biological significance of sequence patterns, injecting explicit sequential awareness and structural priors into such a simple model could be a novel avenue.

Here are three distinct ideas focusing on injecting biological sequence knowledge or enhancing basic feature processing in a compliant manner:

\textbf{Idea 1: ``Contextual Position Embeddings'' (CPE)}

\begin{itemize}
  \item \textbf{Concept:} Instead of relying solely on the MLP to implicitly learn positional information, augment the one-hot encoded RNA input with a small, \emph{context-dependent} positional embedding. This differs from standard fixed (absolute) positional encoding by making the embedding for a position conditional on local sequence context, forcing the model to learn \emph{how} position matters in relation to nearby nucleotides, rather than just \emph{where} it is. This is a subtle yet significant shift.
  \item \textbf{Mechanism:} For each position \texttt{p} in a sequence, a small local window (e.g., \texttt{p-k} to \texttt{p+k}) of the one-hot encoded input is passed through a tiny, shared MLP. The output of this MLP is then linearly projected and added to the original positional embedding (a learned scalar or vector for each position). The key is that the positional embedding itself is modulated by local context, not just an absolute index. This is fundamentally different from a standard fixed PE or learnable PE added directly to input.
  \item \textbf{Compliant Aspect:} The mechanism is internal to the network's \texttt{create\_network} function, applies uniformly across all datasets, and does not use external data or pre-trained models. The MLP backbone would then process this context-enhanced representation.
  \item \textbf{Novelty:} Standard PE is absolute. This ties PE to \emph{local context}, creating a dynamic positional signal relevant to biological sequences.
\end{itemize}

\medskip
\emph{[Idea 2 and 3 here; omitted for brevity]}

\rolelabel{Station:} \textbf{Reflection Tick 2 / 5}

\rolelabel{Agent:}

Continuing with the ideas, let's refine their mechanisms and consider their implications for the simple MLP baseline.

\textbf{Idea 1: ``Contextual Position Embeddings'' (CPE)}

\begin{itemize}
  \item \textbf{Refinement:} The core idea is to make positional embeddings \emph{dependent on local sequence context}.
  \item \textbf{Mechanism Detail:}
    \begin{enumerate}
      \item \textbf{Learned Positional Basis:} A small, fixed-size positional embedding vector (e.g., 16-32 dimensions) is created for each absolute position in the maximum possible sequence length (e.g., 186 for APA). These embeddings are learnable parameters.
      \item \textbf{Context Encoder:} For each position \texttt{p}, extract a local window \texttt{x[p-k : p+k+1]} (one-hot encoded RNA) from the input sequence. This window is fed into a small, shared \texttt{nn.Dense} layer (or a tiny MLP of 2 layers) with GELU activation.
      \item \textbf{Modulation:} The output of the Context Encoder for position \texttt{p} is then \emph{multiplied} (element-wise or via a simple linear layer) with the \emph{learned positional embedding} for position \texttt{p}. This creates a context-modulated positional embedding.
      \item \textbf{Concatenation with Input:} This context-modulated positional embedding is then concatenated to the original one-hot encoded input \texttt{x[p]} for each position. The MLP backbone then processes this augmented input \texttt{(x[p] || CPE[p])}.
    \end{enumerate}
  \item \textbf{Rationale:} Standard positional encodings are fixed regardless of sequence content. This approach forces the model to not just know \emph{where} a nucleotide is, but \emph{how its position is relevant in the context of its neighbors}. For RNA sequences, where motifs and structural elements depend heavily on relative positioning within local windows, this could be highly beneficial. It provides a richer signal to the MLP, which otherwise only sees individual positions in isolation.
\end{itemize}

\medskip
\emph{[Deeper refinement for Idea 2 and 3 here; omitted for brevity]}

\rolelabel{Station:} \textbf{Reflection Tick 3 / 5}

\rolelabel{Agent:}

Having refined the three ideas, I will now assess their novelty, potential impact on the Vanilla MLP baseline, and feasibility within the Stagnation Protocol's ``simple baseline'' and ``new ideas'' constraints.

\medskip
\textbf{Re-evaluating Novelty and Feasibility:}

\begin{itemize}
  \item \textbf{Idea 1: ``Contextual Position Embeddings'' (CPE)}
    \begin{itemize}
      \item \textbf{Novelty:} Standard positional encodings (fixed or learned, absolute) are common. Making them context-dependent (modulating a positional embedding by local sequence content) is genuinely novel. It's a way to fuse local sequence context directly into the positional signal itself.
      \item \textbf{Impact on MLP:} Provides a much richer input representation to the simple MLP, potentially enabling it to capture more complex local relationships than it could from just one-hot input.
      \item \textbf{Feasibility:} Requires careful implementation of the context encoder and modulation. The ``local window'' aspect needs consideration for sequence boundaries (padding) and dynamic sequence lengths. This is definitely feasible within \texttt{create\_network}.
    \end{itemize}
\end{itemize}

\medskip
\emph{[Re-evaluating Idea 2 and 3 here; omitted for brevity]}

\rolelabel{Station:} \textbf{Reflection Tick 4 / 5}

\rolelabel{Agent:}

I'm solidifying the implementation details for ``Contextual Position Embeddings'' (CPE) for the Vanilla MLP Baseline.

\textbf{Implementation Plan for Idea 1 (CPE) - Detailed:}

\medskip
\emph{[Detailed technical code here; omitted for brevity]}

\medskip
This vectorized approach is much more efficient and standard in JAX. I will adopt this in the final \texttt{create\_network} for CPE.

\rolelabel{Station:} \textbf{Reflection Tick 5 / 5}

\rolelabel{Agent:}

\medskip
\emph{[Detailed plan for actions after leaving the Reflection Chamber here; omitted for brevity]}

\end{markdownillustration}
\clearpage

\newpage
\section{Discovered Code}\label{app:code}
\subsection{Circle Packing}\label{app:circle-code}
\begin{codeillustration}{Circle Packing ($n=32$) script submitted by Praxis IV at Tick 343}\label{ill:circle-code}
\begin{lstlisting}[
  style=dualversepy,
  frame=none,           % tcolorbox draws the frame
  numbers=left,
  xleftmargin=1.8em
]
import numpy as np
from scipy.optimize import linprog
from multiprocessing import Pool, cpu_count
import sys

import numpy as np
from scipy.optimize import linprog

def solve_radii_lp(centers, return_duals=False):
    N_CIRCLES = centers.shape[0]
    c = -np.ones(N_CIRCLES)
    A_ub, b_ub = [], []
    num_boundary_cons = 4 * N_CIRCLES
    for i in range(N_CIRCLES):
        cx, cy = centers[i]
        # Simplified constraint creation
        # r_i <= c_x
        row = np.zeros(N_CIRCLES); row[i] = 1; A_ub.append(row); b_ub.append(cx)
        # r_i <= 1 - c_x
        row = np.zeros(N_CIRCLES); row[i] = 1; A_ub.append(row); b_ub.append(1 - cx)
        # r_i <= c_y
        row = np.zeros(N_CIRCLES); row[i] = 1; A_ub.append(row); b_ub.append(cy)
        # r_i <= 1 - c_y
        row = np.zeros(N_CIRCLES); row[i] = 1; A_ub.append(row); b_ub.append(1 - cy)

    indices = np.triu_indices(N_CIRCLES, k=1)
    for i, j in zip(*indices):
        dist = np.linalg.norm(centers[i] - centers[j])
        row = np.zeros(N_CIRCLES); row[i] = 1; row[j] = 1
        A_ub.append(row); b_ub.append(dist)

    res = linprog(c, A_ub=np.array(A_ub), b_ub=np.array(b_ub), bounds=(0, None), method='highs')

    if not res.success:
        return (np.zeros(N_CIRCLES), None) if return_duals else np.zeros(N_CIRCLES)

    if return_duals:
        duals = res.ineqlin.marginals if hasattr(res, 'ineqlin') else None
        return res.x, (duals[:num_boundary_cons], duals[num_boundary_cons:]) if duals is not None else (None, None)
    else:
        return res.x

def get_seed(seed_val):
    rng = np.random.default_rng(seed_val)
    seed_type = seed_val % 3
    if seed_type == 0:
        grid = np.stack(np.meshgrid(np.linspace(0.1,0.9,6),np.linspace(0.1,0.9,6)),-1).reshape(-1,2)[:32]
        return np.clip(grid + (rng.random((32,2))-0.5)*0.08, 0.05, 0.95)
    elif seed_type == 1:
        points = [rng.uniform(0.1, 0.9, 2)]; [points.append(candidates[np.argmax(np.min(np.sum((candidates[:,np.newaxis,:] - np.array(points)[np.newaxis,:,:])**2, axis=-1), axis=1))]) for _ in range(31) for candidates in [rng.uniform(0.1,0.9,(200,2))]]; return np.array(points)
    else:
        rows=[5,6,5,6,5,5]; points=[]; y=0.95
        for i,c in enumerate(rows): x_s=(1.0-(c-1)*0.15)/2; [points.append([x_s+j*0.15,y]) for j in range(c)]; y-=0.15
        return np.array(points)[:32]

def mm_lp_optimize(C_start: np.ndarray, iters: int, delta0: float = 0.015):
    """Iterative MM-LP optimization loop, adapted from Verity I, ID 65."""
    best_C = C_start.copy()
    score_best = np.sum(solve_radii_lp(best_C))
    delta, C_anchor = delta0, C_start.copy()

    for _ in range(iters):
        n = C_anchor.shape[0]
        I, J = np.array([[i,j] for i in range(n) for j in range(i + 1, n)]).T
        V0, d0 = C_anchor[I]-C_anchor[J], np.linalg.norm(C_anchor[I]-C_anchor[J], axis=1)
        d0 = np.maximum(d0, 1e-9); U = (V0.T / d0).T; ux, uy = U[:,0], U[:,1]; x0, y0 = C_anchor[:,0], C_anchor[:,1]

        rows, cols = 4*n+len(I), 3*n
        A_ub, b_ub = np.zeros((rows,cols)), np.zeros(rows)
        r_idx, x_idx, y_idx = range(n), range(n,2*n), range(2*n,3*n)
        
        row = 0
        for i in range(n):
            A_ub[row,r_idx[i]]=1; A_ub[row,x_idx[i]]=-1; row+=1
            A_ub[row,r_idx[i]]=1; A_ub[row,x_idx[i]]=1; b_ub[row]=1.0; row+=1
            A_ub[row,r_idx[i]]=1; A_ub[row,y_idx[i]]=-1; row+=1
            A_ub[row,r_idx[i]]=1; A_ub[row,y_idx[i]]=1; b_ub[row]=1.0; row+=1
        for k in range(len(I)):
            i,j=I[k],J[k]; A_ub[row,r_idx[i]]=1; A_ub[row,r_idx[j]]=1
            A_ub[row,x_idx[i]]=-ux[k]; A_ub[row,y_idx[i]]=-uy[k]; A_ub[row,x_idx[j]]=ux[k]; A_ub[row,y_idx[j]]=uy[k]
            b_ub[row]=d0[k]-ux[k]*x0[i]-uy[k]*y0[i]+ux[k]*x0[j]+uy[k]*y0[j]; row+=1
        
        c=np.zeros(cols); c[:n]=-1.0
        bounds=[(0,0.5)]*n + [(max(1e-6,x0[i]-delta),min(1-1e-6,x0[i]+delta)) for i in range(n)] + [(max(1e-6,y0[i]-delta),min(1-1e-6,y0[i]+delta)) for i in range(n)]
        res = linprog(c,A_ub=A_ub,b_ub=b_ub,bounds=bounds,method="highs")

        if not res.success: delta*=0.6; continue
        
        C_lin = np.stack([res.x[n:2*n], res.x[2*n:3*n]], axis=1)
        score_exact = np.sum(solve_radii_lp(C_lin))
        if score_exact > score_best:
            best_C, score_best = C_lin.copy(), score_exact
            C_anchor = C_lin.copy(); delta=min(0.05, 1.2*delta)
        else: delta*=0.8
    
    final_radii = solve_radii_lp(best_C)
    return best_C, final_radii, score_best

def worker_function(args):
    seed_val, iters, delta = args
    initial_centers = get_seed(seed_val)
    _, _, score = mm_lp_optimize(initial_centers, iters, delta)
    # Note: We only need the score for prospecting. The final centers are not needed yet.
    return score, initial_centers # Return centers to avoid re-seeding

def construct_packing() -> tuple:
    N_PROSPECT, N_ELITE = 1024, 32
    PROSPECT_ITER, REFINE_ITER = 10, 20

    with Pool(processes=cpu_count()) as pool:
        prospect_results = pool.map(worker_function, [(i, PROSPECT_ITER, 0.015) for i in range(N_PROSPECT)])
    
    prospect_results.sort(key=lambda x: x[0], reverse=True)
    elite_centers = [res[1] for res in prospect_results[:N_ELITE]]

    with Pool(processes=cpu_count()) as pool:
        refine_results = pool.map(lambda centers: mm_lp_optimize(centers, REFINE_ITER, 0.015), elite_centers)
    
    best_centers, best_radii, best_score = None, None, -1.0
    for centers, radii, score in refine_results:
        if score > best_score:
            best_centers, best_radii, best_score = centers, radii, score
            
    if best_centers is None: return get_seed(0), np.full((32,), 0.05)
    
    return best_centers, best_radii * (1.0 - 1e-9)
\end{lstlisting}
\end{codeillustration}

\subsection{Batch Integration of Single Cell RNA Sequencing Data}\label{app:bi-code}
\begin{codeillustration}{Batch integration script submitted by Telos I at Tick 1467}\label{ill:bi-code}
\begin{lstlisting}[
  style=dualversepy,
  frame=none,           % tcolorbox draws the frame
  numbers=left,
  xleftmargin=1.8em
]
## telos/bbsg_density_adaptive.py
import numpy as np
import scipy.sparse as sp
from sklearn.neighbors import NearestNeighbors

def _symmetrize_binary_with_distances(rows, cols, dists, n):
    A = sp.coo_matrix((np.ones_like(dists, dtype=np.float32), (rows, cols)), shape=(n, n)).tocsr()
    D = sp.coo_matrix((dists.astype(np.float32), (rows, cols)), shape=(n, n)).tocsr()
    A_sym = A.maximum(A.T)
    D_sym = D.maximum(D.T)
    A_sym.eliminate_zeros()
    D_sym.eliminate_zeros()
    return A_sym.tocsr().astype(np.float32), D_sym.tocsr().astype(np.float32)

def build_density_adaptive_bbsg(Zcorr, batches, k_total=48, metric='cosine', delta=0.15, k_density=30, rng_seed=0):
    """
    Density-adaptive BBSG:
      - Per-cell cross-batch fraction f_cross = base_cross +- delta scaled by local density (dense => more cross-batch)
      - Within-batch and cross-batch quotas allocated per cell, neighbor selection via per-batch kNN.
      - Returns binary connectivities and true distances (CSR).
    Args:
      Zcorr: (n_cells, d) residualized PC array
      batches: (n_cells,) categorical array of batch labels
      k_total: total neighbors per cell
      delta: max +- modulation of cross-batch fraction (e.g., 0.15 -> +-15%)
      k_density: neighbor rank for local density proxy
    """
    n = Zcorr.shape[0]
    cats = np.unique(batches)
    B_other = max(len(cats) - 1, 1)

    # Local density proxy (k-th NN distance, all batches)
    nn_all = NearestNeighbors(n_neighbors=k_density + 1, metric=metric, algorithm='brute')
    nn_all.fit(Zcorr)
    d_all, idx_all = nn_all.kneighbors(Zcorr, return_distance=True)
    d_k = d_all[:, -1]
    d_min, d_max = float(np.min(d_k)), float(np.max(d_k) + 1e-8)
    d_norm = (d_k - d_min) / (d_max - d_min + 1e-8)        # 0 dense ... 1 sparse
    inv_dense = 1.0 - d_norm
    base_cross = (len(cats) - 1) / float(len(cats))        # e.g., 0.75 for 4 batches
    mix_delta = (inv_dense - 0.5) * 2.0 * float(delta)     # [-delta, +delta]
    f_cross = np.clip(base_cross + mix_delta, 0.60, 0.90)

    # Pre-build per-batch NN indices and distances
    max_cross_per = int(np.ceil(0.9 * k_total / B_other))
    max_within = int(np.ceil((1.0 - 0.60) * k_total))
    rng = np.random.RandomState(rng_seed)

    batch_to_indices = {}
    nn_models = {}
    for j, c in enumerate(cats):
        mask = (batches == c)
        batch_to_indices[j] = np.where(mask)[0]
        nn = NearestNeighbors(n_neighbors=max(max_within, max_cross_per) + 1, metric=metric, algorithm='brute')
        nn.fit(Zcorr[mask])
        nn_models[j] = nn

    nn_dists = {}
    nn_idx_local = {}
    for j in range(len(cats)):
        d, ii = nn_models[j].kneighbors(Zcorr, return_distance=True)
        nn_dists[j] = d
        nn_idx_local[j] = ii

    rows, cols, dvals = [], [], []
    batch_codes = {val: idx for idx, val in enumerate(cats)}
    bc = np.array([batch_codes[b] for b in batches], dtype=int)

    for i in range(n):
        bi = int(bc[i])
        q_cross_total = int(round(f_cross[i] * k_total))
        q_within = max(0, min(k_total, int(round(k_total - q_cross_total))))
        q_per_other = q_cross_total // B_other
        rem = q_cross_total - q_per_other * B_other

        # within-batch
        d_i = nn_dists[bi][i]
        ii = nn_idx_local[bi][i]
        start = 1 if d_i[0] == 0.0 else 0
        sel = min(q_within, d_i.shape[0] - start)
        if sel > 0:
            cols.extend(batch_to_indices[bi][ii[start:start + sel]])
            rows.extend([i] * sel)
            dvals.extend(d_i[start:start + sel])

        # other-batch allocation (distribute remainder to closest batches)
        other_batches = [j for j in range(len(cats)) if j != bi]
        batch_scores = [(j, nn_dists[j][i][0]) for j in other_batches]
        batch_scores.sort(key=lambda t: t[1])
        q_map = {j: q_per_other for j in other_batches}
        for k in range(rem):
            q_map[batch_scores[k % len(other_batches)][0]] += 1

        for j in other_batches:
            d_ij = nn_dists[j][i]
            ii_j = nn_idx_local[j][i]
            sel_j = min(q_map[j], d_ij.shape[0])
            if sel_j > 0:
                cols.extend(batch_to_indices[j][ii_j[:sel_j]])
                rows.extend([i] * sel_j)
                dvals.extend(d_ij[:sel_j])

    rows = np.asarray(rows, dtype=np.int32)
    cols = np.asarray(cols, dtype=np.int32)
    dvals = np.asarray(dvals, dtype=np.float32)
    C_sym, D_sym = _symmetrize_binary_with_distances(rows, cols, dvals, n)
    return C_sym, D_sym

## main.py 
# To adapt to OpenProblems v2.0, see their docs for required output metadata such as `uns` fields (dataset_id, normalization_id, method_id).
import sys, numpy as np, scanpy as sc, anndata as ad
from scipy.sparse import issparse

ALPHA_VAR_GRAPH = 0.6
N_HVGS = 1500; K_TOTAL = 48; KDENSITY = 20; DELTA = 0.10; METRIC = 'cosine'; EMB_N_NEIGHBORS = 15; PCA_EMB = 60; ALPHA_VAR_EMB = 0.5

def eliminate_batch_effect_fn(adata: ad.AnnData) -> ad.AnnData:
    sc.pp.normalize_total(adata, 1e4); sc.pp.log1p(adata)
    sc.pp.highly_variable_genes(adata, flavor='seurat_v3', batch_key='batch', n_top_genes=N_HVGS, inplace=True); hv = adata.var['highly_variable'].to_numpy(); batches = np.asarray(adata.obs['batch'].astype('category').values)
    Xh_emb = adata.X[:, hv]; Xd_emb = (Xh_emb.A if issparse(Xh_emb) else Xh_emb).astype(np.float32, copy=False); var_g_emb = Xd_emb.var(axis=0, ddof=1) + 1e-8; wv_emb = np.power(var_g_emb, -0.5 * ALPHA_VAR_EMB).astype(np.float32); Xw_emb = Xd_emb * wv_emb; ad_emb = ad.AnnData(Xw_emb, obs=adata.obs[['batch']].copy()); sc.pp.pca(ad_emb, n_comps=PCA_EMB, random_state=0); ad_pc = ad.AnnData(ad_emb.obsm['X_pca'].copy(), obs=adata.obs[['batch']].copy()); sc.pp.combat(ad_pc, key='batch'); adata.obsm['X_emb'] = np.asarray(ad_pc.X, dtype=np.float32); sc.pp.neighbors(adata, use_rep='X_emb', n_neighbors=EMB_N_NEIGHBORS, metric=METRIC)
    Xh_graph = adata.X[:, hv]; Xd_graph = (Xh_graph.A if issparse(Xh_graph) else Xh_graph).astype(np.float32, copy=False)
    var_g_graph = Xd_graph.var(axis=0, ddof=1) + 1e-8; wv_graph = np.power(var_g_graph, -0.5 * ALPHA_VAR_GRAPH).astype(np.float32); Xw_graph = Xd_graph * wv_graph
    adata_proc = ad.AnnData(Xw_graph); sc.pp.pca(adata_proc, n_comps=50, random_state=0); Zcorr = adata_proc.obsm['X_pca']
    Cg, Dg = build_density_adaptive_bbsg(Zcorr.astype(np.float32), batches=batches, k_total=K_TOTAL, metric=METRIC, delta=DELTA, k_density=KDENSITY, rng_seed=0)
    adata.obsp['connectivities'] = Cg; adata.obsp['distances'] = Dg
    return adata
\end{lstlisting}
\end{codeillustration}

\subsection{Prediction of Neural Activity on ZAPBench}\label{app:zapbench-code}
\begin{codeillustration}{ZAPBench script submitted by Episteme II at Tick 358}\label{ill:zapbench-code}
\begin{lstlisting}[
  style=dualversepy,
  frame=none,           % tcolorbox draws the frame
  numbers=left,
  xleftmargin=1.8em
]
import jax
import jax.numpy as jnp
import flax.linen as nn
from jax import random
import optax

def mae_with_temporal_curvature(predictions, targets, params, x, lam=1e-4):
    mae = jnp.mean(jnp.abs(predictions - targets))
    curv = predictions[:, 2:, :] - 2.0 * predictions[:, 1:-1, :] + predictions[:, :-2, :]
    curv_pen = jnp.mean(curv * curv)
    return mae + lam * curv_pen

class HyperGain(nn.Module):
    output_dim: int
    @nn.compact
    def __call__(self, x):
        h = nn.Dense(32)(x); h = nn.relu(h)
        # Initialize gain to be close to 1
        return nn.Dense(self.output_dim, kernel_init=nn.initializers.variance_scaling(0.02, 'fan_in', 'truncated_normal'))(h) + 1.0

class HyperResidualCopyHead(nn.Module):
    num_neurons: int
    embedding_dim: int
    drop: float = 0.05
    @nn.compact
    def __call__(self, x, training: bool = False):
        B, T_in, N = x.shape
        x_flat = x.transpose(0, 2, 1).reshape(B * N, T_in)

        neuron_indices = jnp.arange(self.num_neurons)
        neuron_embeddings = nn.Embed(num_embeddings=self.num_neurons, features=self.embedding_dim)(neuron_indices)
        tiled_embeddings = jnp.tile(neuron_embeddings, (B, 1))

        gain1 = HyperGain(64, name="hyper_gain_residual")(tiled_embeddings)

        h = nn.Dense(64)(x_flat) * gain1 # Multiplicative conditioning
        h = nn.relu(h)
        h = nn.Dropout(rate=self.drop)(h, deterministic=not training)
        h = nn.Dense(32)(h)
        residual = h.reshape(B, N, 32).transpose(0, 2, 1)

        out = jnp.zeros((B, 32, N), dtype=residual.dtype)
        out = out.at[:, :T_in, :].set(x)
        return out + residual

class FourierForecasterLN_Ramp(nn.Module):
    rank_k: int; proj_rank: int; hidden_size: int; drop: float; num_neurons: int; embedding_dim: int;
    output_horizon: int = 32; learn_gamma: bool = True
    @nn.compact
    def __call__(self, x, training: bool = False):
        # ... (SOTA Fourier branch code is identical and omitted for brevity) ...
        B, T_in, N = x.shape; k, p = self.rank_k, self.proj_rank
        U = self.param('U', nn.initializers.lecun_normal(), (N, p)); V = self.param('V', nn.initializers.lecun_normal(), (k, p))
        U_eff = U / (jnp.linalg.norm(U, axis=0, keepdims=True) + 1e-6)
        x_fft = jnp.fft.rfft(x, axis=1); x_fft_bn = jnp.transpose(x_fft, (0, 2, 1))
        f_in_fft = jnp.einsum('bnf,np->bfp', x_fft_bn, U_eff) @ V.T
        f_ri = jnp.concatenate([f_in_fft.real, f_in_fft.imag], axis=-1); f_flat = f_ri.reshape(B, -1)
        h = nn.Dense(self.hidden_size)(f_flat); h = nn.relu(h); h = nn.Dropout(rate=self.drop)(h, deterministic=not training)
        h = nn.Dense(self.hidden_size)(h); h = nn.relu(h); h = nn.Dropout(rate=self.drop)(h, deterministic=not training)
        F_out = self.output_horizon // 2 + 1; out_flat = nn.Dense(F_out * k * 2)(h)
        out_ri = out_flat.reshape(B, F_out, k, 2); f_fft = out_ri[..., 0] + 1j * out_ri[..., 1]
        f_fft_n = nn.LayerNorm()(f_fft.real) + 1j * nn.LayerNorm()(f_fft.imag)
        y_fft = (f_fft_n @ V) @ U_eff.T; y_fourier = jnp.fft.irfft(y_fft, n=self.output_horizon, axis=1)
        gamma_raw = self.param('gamma_raw', nn.initializers.zeros, (self.output_horizon,)); gamma = nn.sigmoid(gamma_raw)[None, :, None]
        y_fourier = gamma * y_fourier
        y_copy = HyperResidualCopyHead(num_neurons=self.num_neurons, embedding_dim=self.embedding_dim, drop=self.drop)(x, training=training)
        return y_copy + y_fourier

BASE_SEED = 42
BATCH_SIZE = 8

def _define_hyperparameters():
    return {
        'learning_rate': 9e-4, 'rank_k': 320, 'proj_rank': 36, 'hidden_size': 160,
        'drop': 0.05, 'learn_gamma': True, 'embedding_dim': 8, 'num_neurons': 71721
    }

class Wrapper:
    def __init__(self, hparams):
        model_params = {k:v for k,v in hparams.items() if k not in ['learning_rate']}
        self.model = FourierForecasterLN_Ramp(**model_params)
        self.needs_rng = True; self.mutable = []
    def init(self, rng_key, dummy_input):
        rng_params, rng_dropout = random.split(rng_key)
        return self.model.init({'params': rng_params, 'dropout': rng_dropout}, dummy_input, training=True)
    def apply(self, params, x, training=False, mutable=None, rngs=None):
        return self.model.apply(params, x, training=training, rngs=(rngs or {}))

def create_network(hparams): return Wrapper(hparams)
def compute_loss(p, t, pa, x): return mae_with_temporal_curvature(p, t, pa, x, lam=1e-4)
def create_optimizer(lr):
    return optax.chain(optax.clip_by_global_norm(1.0), optax.adam(learning_rate=lr))
\end{lstlisting}
\end{codeillustration}

\subsection{Reinforcement Learning on Sokoban}\label{app:sokoban-code}
\begin{codeillustration}{Sokoban script submitted by Zephyr II at Tick 552}\label{ill:sokoban-code}
\begin{lstlisting}[
  style=dualversepy,
  frame=none,           % tcolorbox draws the frame
  numbers=left,
  xleftmargin=1.8em
]
import jax, jax.numpy as jnp
import flax.linen as nn
import optax
from ray import tune
from typing import Dict, Any, Optional

class BottleneckDilatedBlock(nn.Module):
    features: int; bottleneck_ratio: float = 0.28; dilation: int = 6
    @nn.compact
    def __call__(self, x):
        mid = max(1, int(self.features * self.bottleneck_ratio))
        y = nn.Conv(mid,(1,1),padding='SAME',kernel_init=nn.initializers.xavier_uniform())(x); y = nn.relu(y)
        y = nn.Conv(mid,(3,3),padding='SAME',kernel_dilation=(self.dilation,self.dilation),
                    kernel_init=nn.initializers.xavier_uniform())(y); y = nn.relu(y)
        y = nn.Conv(self.features,(1,1),padding='SAME',kernel_init=nn.initializers.xavier_uniform())(y)
        return nn.relu(x + y)

class ConvLSTMCellLN(nn.Module):
    features: int
    @nn.compact
    def __call__(self, carry, x):
        h, c = carry
        gates = nn.Conv(4*self.features,(3,3),padding='SAME',
                        kernel_init=nn.initializers.xavier_uniform())(jnp.concatenate([x,h],axis=-1))
        gates = nn.LayerNorm(epsilon=1e-5)(gates)
        i,f,g,o = jnp.split(gates, 4, axis=-1)
        i = nn.sigmoid(i); f = nn.sigmoid(f); o = nn.sigmoid(o); g = jnp.tanh(g)
        c = f*c + i*g
        h = o*jnp.tanh(c)
        return (h,c), h

class NetRIN_AttnGap(nn.Module):
    rin_alpha: float = 0.33
    cnn_features_1: int = 32; cnn_features_2: int = 64
    convlstm_features: int = 64; dilation: int = 6; bottleneck_ratio: float = 0.28
    head_hidden: int = 48; steps: int = 4
    @nn.compact
    def __call__(self, x: jnp.ndarray, done: jnp.ndarray, rnn_state: Optional[Dict]=None):
        B = x.shape[0]
        x = nn.Conv(self.cnn_features_1,(3,3),padding='SAME',kernel_init=nn.initializers.xavier_uniform())(x); x = nn.relu(x)
        x = BottleneckDilatedBlock(self.cnn_features_1, self.bottleneck_ratio, self.dilation)(x)
        x = nn.Conv(self.cnn_features_2,(3,3),padding='SAME',kernel_init=nn.initializers.xavier_uniform())(x); x = nn.relu(x)
        x = BottleneckDilatedBlock(self.cnn_features_2, self.bottleneck_ratio, self.dilation)(x)
        x_ln = nn.LayerNorm(epsilon=1e-5)(x)
        x = x + self.rin_alpha * (x_ln - x)
        if rnn_state is None:
            zeros = lambda: jnp.zeros((B,8,8,self.convlstm_features), dtype=x.dtype)
            rnn_state = {'h': zeros(), 'c': zeros()}
        if done is not None:
            m = (1.0 - done.reshape(B,1,1,1).astype(x.dtype))
            rnn_state = {'h': rnn_state['h']*m, 'c': rnn_state['c']*m}
        h, c = rnn_state['h'], rnn_state['c']
        for _ in range(self.steps):
            (h,c), _ = ConvLSTMCellLN(self.convlstm_features)((h,c), x)
        B_,H,W,C = h.shape
        attn_logits = nn.Conv(1,(1,1),padding='SAME',kernel_init=nn.initializers.xavier_uniform())(h).reshape((B_,H*W))
        attn_w = nn.softmax(attn_logits, axis=-1).reshape((B_,H*W,1))
        tokens = h.reshape((B_,H*W,C))
        attn_vec = (attn_w * tokens).sum(axis=1)
        gap_vec = tokens.mean(axis=1)
        z = jnp.concatenate([attn_vec, gap_vec], axis=-1)
        shared = nn.Dense(256, kernel_init=nn.initializers.xavier_uniform())(z); shared = nn.relu(shared)
        ph = nn.Dense(self.head_hidden, kernel_init=nn.initializers.xavier_uniform())(shared); ph = nn.relu(ph)
        policy = nn.Dense(4, kernel_init=nn.initializers.xavier_uniform())(ph)
        vh = nn.Dense(self.head_hidden, kernel_init=nn.initializers.xavier_uniform())(shared); vh = nn.relu(vh)
        value = nn.Dense(1, kernel_init=nn.initializers.xavier_uniform())(vh).squeeze(-1)
        return policy, value, {'h': h, 'c': c}

BASE_SEED = 42

def _define_hyperparameters():
    return {
        'learning_rate': tune.choice([3.5e-4]),
        'entropy_coef': tune.choice([0.009, 0.010, 0.011]),
        'value_loss_coef': tune.choice([0.55]),
        'cnn_features_1': tune.choice([32]),
        'cnn_features_2': tune.choice([64]),
        'convlstm_features': tune.choice([64]),
        'dilation': tune.choice([6]),
        'bottleneck_ratio': tune.choice([0.28]),
        'head_hidden': tune.choice([48]),
        'steps': tune.choice([4]),
    }

def create_network(hparams: Dict[str, Any]):
    return NetRIN_AttnGap(
        rin_alpha=0.33,
        cnn_features_1=int(hparams['cnn_features_1']),
        cnn_features_2=int(hparams['cnn_features_2']),
        convlstm_features=int(hparams['convlstm_features']),
        dilation=int(hparams['dilation']),
        bottleneck_ratio=float(hparams['bottleneck_ratio']),
        head_hidden=int(hparams['head_hidden']),
        steps=int(hparams['steps']),
    )

def create_optimizer(learning_rate: float = 3.5e-4):
    return optax.chain(optax.clip_by_global_norm(1.0), optax.adam(learning_rate))
\end{lstlisting}
\end{codeillustration}

\subsection{RNA Modeling}\label{app:rna-code}
\begin{codeillustration}{RNA modelling script submitted by Quaero I at Tick 577}\label{ill:rna-code}
\begin{lstlisting}[
  style=dualversepy,
  frame=none,           % tcolorbox draws the frame
  numbers=left,
  xleftmargin=1.8em
]
import sys
import jax
import jax.numpy as jnp
import flax.linen as nn
import optax
from typing import Any, Dict, Sequence
from typing import Any

# --- Existing Proven Components ---

class ResidualDSConvBlock(nn.Module):
    d_model: int; kernel_size: int; dilation: int; dropout_rate: float
    @nn.compact
    def __call__(self, x, deterministic=True):
        residual = x
        x = nn.LayerNorm()(x)
        x = nn.Conv(features=self.d_model, kernel_size=(self.kernel_size,), padding='SAME', feature_group_count=self.d_model, kernel_dilation=(self.dilation,))(x)
        x = nn.Dense(self.d_model)(x)
        x = nn.gelu(x)
        x = nn.Dropout(rate=self.dropout_rate)(x, deterministic=deterministic)
        return x + residual

class NetworkWrapper:
    def __init__(self, module):
        self.module = module
    def apply(self, params, *args, rng_key=None, **kwargs):
        if rng_key is not None:
            return self.module.apply(params, *args, rngs={'dropout': rng_key}, **kwargs)
        else:
            return self.module.apply(params, *args, **kwargs)
    def init(self, *args, **kwargs):
        return self.module.init(*args, **kwargs)

# --- New CPE Components (inspired by Veritas III, Archive #33) ---

class ContextEncoder(nn.Module):
    output_dim: int
    @nn.compact
    def __call__(self, local_context_window):
        flat_context = local_context_window.reshape(local_context_window.shape[0], -1)
        encoded_context = nn.Dense(self.output_dim, name='context_dense')(flat_context)
        return nn.gelu(encoded_context)

class CPE_Generator(nn.Module):
    pos_embedding_dim: int = 16
    context_window_size: int = 3
    max_seq_len: int = 1200
    d_input: int = 4
    @nn.compact
    def __call__(self, x_raw):
        batch_size, seq_len, _ = x_raw.shape
        pos_table = self.param('pos_embeddings_table', nn.initializers.normal(stddev=0.01), (self.max_seq_len, self.pos_embedding_dim))
        pad_width = self.context_window_size // 2
        padded_x = jnp.pad(x_raw, ((0,0), (pad_width, pad_width), (0,0)), mode='constant')
        windows = jnp.stack([padded_x[:, i:i+self.context_window_size, :].reshape(batch_size, -1) for i in range(seq_len)], axis=1)
        encoder = ContextEncoder(output_dim=self.pos_embedding_dim, name='context_encoder')
        encoded_contexts = encoder(windows.reshape(-1, self.context_window_size * self.d_input)).reshape(batch_size, seq_len, -1)
        pos_embeddings = jnp.tile(pos_table[:seq_len][None, :, :], (batch_size, 1, 1))
        return pos_embeddings * encoded_contexts

class DeepCPE_Gated_ResidualDSConvBlock(nn.Module):
    d_model: int; kernel_size: int; dilation: int; dropout_rate: float
    cpe_signals: Any
    gate_intermediate_dim: int = 64
    @nn.compact
    def __call__(self, x, deterministic=True):
        residual = x
        h = nn.LayerNorm()(x)
        dw_out = nn.Conv(features=self.d_model, kernel_size=(self.kernel_size,), feature_group_count=self.d_model, padding='SAME', kernel_dilation=(self.dilation,))(h)
        h_conv_output = nn.Dense(self.d_model)(nn.gelu(dw_out))

        cpe_projected = nn.gelu(nn.Dense(self.d_model, name='cpe_proj_for_gate')(self.cpe_signals))
        combined = jnp.concatenate([h_conv_output, cpe_projected], axis=-1)
        gate_input = nn.gelu(nn.Dense(self.gate_intermediate_dim, name='gate_dense_1')(combined))
        gate_factor = nn.sigmoid(nn.Dense(self.d_model, name='gate_dense_2')(gate_input))

        modulated_h = h_conv_output * gate_factor
        modulated_h = nn.Dropout(rate=self.dropout_rate)(modulated_h, deterministic=deterministic)
        return residual + modulated_h

# --- Standard Prediction Head Modules ---
class RegressionHead(nn.Module):
    d_output: int; dense_dim: int; dropout_rate: float
    @nn.compact
    def __call__(self, x, deterministic=True):
        norm = jnp.linalg.norm(x, ord=2, axis=-1, keepdims=True)
        x = x / (norm + 1e-6)
        z = nn.Dense(self.dense_dim)(x); z = nn.gelu(z)
        z = nn.Dropout(rate=self.dropout_rate)(z, deterministic=deterministic)
        return nn.Dense(self.d_output)(z).squeeze(-1)

class GenericHead(nn.Module):
    d_output: int; dense_dim: int; dropout_rate: float
    @nn.compact
    def __call__(self, x, deterministic=True):
        z = nn.Dense(self.dense_dim)(x); z = nn.gelu(z)
        z = nn.Dropout(rate=self.dropout_rate)(z, deterministic=deterministic)
        return nn.Dense(self.d_output)(z)

# --- Main Network with CPE ---
class QuaeroCPE_SynthesisNet(nn.Module):
    d_output: int; task_type: str; d_model_common: int; ds_num_blocks: int
    ds_dilations: Sequence[int]; ds_kernel_size: int; motif_kernel_size: int
    dropout_rate: float; head_intermediate_dim: int

    @nn.compact
    def __call__(self, x, deterministic=True, **kwargs):
        # 1. Generate CPE signals once from the raw input
        cpe_signals = CPE_Generator(name='cpe_generator')(x)

        # 2. Initial Projection
        h = nn.Dense(self.d_model_common)(x); h = nn.gelu(h)

        # 3. CPE-Gated Residual Backbone
        for i in range(self.ds_num_blocks):
            h = DeepCPE_Gated_ResidualDSConvBlock(
                d_model=self.d_model_common, kernel_size=self.ds_kernel_size,
                dilation=(self.ds_dilations[i]), dropout_rate=self.dropout_rate,
                cpe_signals=cpe_signals, name=f"cpe_gated_block_{i}"
            )(h, deterministic=deterministic)
        cnn_features = h

        # 4. Dual-Path Pooling (Unchanged from ID 888)
        main_path = jnp.concatenate([jnp.mean(cnn_features, axis=1), jnp.max(cnn_features, axis=1)], axis=-1)
        motif_f = nn.gelu(nn.Conv(self.d_model_common, (self.motif_kernel_size,), padding='SAME')(cnn_features))
        specialized_path = jnp.max(motif_f, axis=1)
        fused = jnp.concatenate([main_path, specialized_path], axis=-1)
        fused = nn.LayerNorm()(fused)
        fused = nn.Dropout(rate=self.dropout_rate)(fused, deterministic=deterministic)

        # 5. Task-Specific Heads (Unchanged from ID 888)
        head_args = {'d_output': self.d_output, 'dense_dim': self.head_intermediate_dim, 'dropout_rate': self.dropout_rate}
        if self.task_type == "regression": return RegressionHead(**head_args)(fused, deterministic=deterministic)
        else: return GenericHead(**head_args)(fused, deterministic=deterministic)

def _define_hyperparameters():
    # Use the optimal hyperparameter set from ID 888
    return {
        'learning_rate': 0.001, 'd_model_common': 256, 'ds_num_blocks': 5,
        'ds_dilations': (1, 2, 4, 8, 16), 'ds_kernel_size': 7, 'motif_kernel_size': 5,
        'dropout_rate': 0.1, 'head_intermediate_dim': 128, 'weight_decay': 0.02,
    }

def create_network(hparams: Dict[str, Any]):
    network_params = {k: hparams.get(k) for k in _define_hyperparameters() if k not in ['learning_rate', 'weight_decay']}
    network_params.update({'d_output': hparams.get('d_output'), 'task_type': hparams.get('task_type')})
    return NetworkWrapper(QuaeroCPE_SynthesisNet(**network_params))

def create_optimizer(learning_rate: float = 0.001, weight_decay: float = 0.02):
    return optax.chain(
        optax.clip_by_global_norm(0.75),
        optax.adamw(learning_rate=learning_rate, weight_decay=weight_decay)
    )
\end{lstlisting}
\end{codeillustration}

\end{document}